%% file: main.tex
\documentclass[lettersize,journal]{IEEEtran}

\usepackage[nocompress]{cite}

\usepackage{colortbl}
\usepackage{amsthm}
\usepackage{amsmath,amsfonts}
\usepackage{algorithmic}
\usepackage{algorithm}
\newtheorem{theorem}{Theorem}

\usepackage{colortbl}  
\usepackage{xcolor}
\usepackage{array}   
\usepackage{makecell}

\usepackage{array}
\usepackage{booktabs}
\usepackage{algorithmic}
\usepackage{arydshln}
\usepackage{ctable}

\usepackage{color}
\usepackage{bm}
\usepackage{makecell}
\usepackage{ragged2e}
\usepackage{balance}
\usepackage{multirow}
\usepackage{booktabs}

\usepackage{amsthm}
\usepackage{amssymb}

\newtheorem{proof}{Proof}
\usepackage{array}
\usepackage{textcomp}
\usepackage{stfloats}
\usepackage{url}
\usepackage{verbatim}
\usepackage{graphicx}
\usepackage[labelformat=simple]{subcaption}

\newcommand{\myparagraph}[1]{\vspace{3.0pt}\noindent{\bf #1}}

\usepackage{amsmath}

\definecolor{mygray}{gray}{.9}

\begin{document}

\title{Inference-Time Rule Eraser: Fair Recognition via Distilling and Removing Biased Rules}


\author{Yi~Zhang, Dongyuan~Lu,~\IEEEmembership{Member,~IEEE,}
and Jitao~Sang,~\IEEEmembership{Member,~IEEE} 
\IEEEcompsocitemizethanks{
  \IEEEcompsocthanksitem Yi Zhang and Jitao Sang are with the School of Computer and Information Technology \& Beijing Key Lab of Traffic Data Analysis and Mining, Beijing Jiaotong University, China. Jitao Sang is also with also with Peng Cheng Laboratory, Shenzhen, China. Email: \{yi.zhang, jtsang\}bjtu.edu.cn. 
  \IEEEcompsocthanksitem Dongyuan Lu is with the School of Information Technology \& Management, University of International Business and Economics, Beijing, China (E-mail: ludy@uibe.edu.cn)
  \IEEEcompsocthanksitem Corresponding author: Dongyuan Lu.
}}

\markboth{Journal of \LaTeX\ Class Files,~Vol.~14, No.~8, August~2021}%
{Shell \MakeLowercase{\textit{et al.}}: A Sample Article Using IEEEtran.cls for IEEE Journals}


\maketitle



\begin{abstract}
Machine learning models often make predictions based on biased features such as gender, race, and other social attributes, posing significant fairness risks, especially in societal applications, such as hiring, banking, and criminal justice. Traditional approaches to addressing this issue involve retraining or fine-tuning neural networks with fairness-aware optimization objectives. However, these methods can be impractical due to significant computational resources, complex industrial tests, and the associated CO2 footprint. Additionally, regular users often fail to fine-tune models because they lack access to model parameters. In this paper, we introduce the Inference-Time Rule Eraser (Eraser), a novel method designed to address fairness concerns by removing biased decision-making rules from deployed models during inference without altering model weights. We begin by establishing a theoretical foundation for modifying model outputs to eliminate biased rules through Bayesian analysis. Next, we present a specific implementation of Eraser that involves two stages: (1) distilling the biased rules from the deployed model into an additional patch model, and (2) removing these biased rules from the output of the deployed model during inference. Extensive experiments validate the effectiveness of our approach, showcasing its superior performance in addressing fairness concerns in AI systems.
\end{abstract}

\begin{IEEEkeywords}
Fairness machine learning, Debiasing, Model rules editing.
\end{IEEEkeywords}


\section{Introduction}\label{sec1}
\IEEEPARstart{A}{Artificial} intelligence (AI) systems have become increasingly prevalent in the real world and are widely deployed in many high-stakes applications such as face recognition and recruitment. 
However, these systems often inherit biases present in their training data, leading to unfair treatment based on protected attributes such as gender and race~\cite{dastin2022amazon,raghavan2020mitigating}. For example, the popular COMPAS algorithm for recidivism prediction exhibited bias against Black inmates and made unfair sentencing decisions~\cite{dressel2018accuracy}. Similarly, Microsoft's face recognition system showed gender bias in classifying face attributes~\cite{buolamwini2018gender}. The prevalence of AI unfairness has surged in recent years, emphasizing the need to prioritize fairness alongside accuracy.

Mainstream solutions to fairness issues often involve incorporating fairness-aware constraints into training algorithms, such as fair contrastive learning~\cite{park2022fair} and adversarial debiasing~\cite{wang2019balanced}. However, these approaches typically require retraining or fine-tuning the deployed model, which can be impractical in real-world scenarios due to substantial computational costs and the associated CO2 footprint. Moreover, machine learning models are often deployed as black-box services, making it difficult for regular users and third parties to access or modify the model's parameters. Alternative approaches, such as optimizing the transformation function attached to the model output~\cite{pleiss2017fairness, hardt2016equality}, have been explored to debias models without retraining. However, these methods require access to additional annotation information of the test sample during inference and often lead to undesirable accuracy reduction, posing challenges in real-world applications. Other techniques~\cite{wang2022fairness,zhang2022fairness} involve introducing additional pixel perturbations using separate perturbation networks, but this approach requires the deployed model to be a white box for training the perturbation network. Consequently, developing practical debiasing methods for deployed black-box models remains a challenging and urgent problem.

In this paper, we aim to address the problem of debiasing deployed models in practical scenarios, focusing on situations where only the model's output can be obtained.
Ideal bias mitigation for the deployed model should meet two objectives: (1) eliminate the biased rule, which is unintentionally learned due to biased training data, such as making predictions based on protected attributes like gender or race and (2) preserve the target rule, which aligns with the model's intended goals, such as predicting candidates' suitability based on their skills in a recruitment model. Hence here come our research questions: \emph{Can we exclusively remove biased rules in deployed models when we can only obtain the model’s output? If so, why and how can we remove them?}

We present the answer is ``Yes'', by looking into model unfairness from the perspective of Bayesian analysis. Specifically, we initially employ Bayesian theory to separately explain the conditional probability $p(y|\mathbf{x})$ (\emph{i.e.}, the probability of the model prediction input $\mathbf{x}$ is $y$) corresponding to models containing mixed rules (both target and biased rules) and models containing only target rules. Based on this understanding,
we then derive the theoretical foundation of the proposed Inference-Time Rule Eraser (Eraser) for removing biased rules: we need only to subtract the response of biased rules (i.e., $p(y|b)$) from the original output of the deployed model to get the repaired fair decision. This method of removing biased rules solely from model output ensures that we do not need to access or modify the model parameters.

\begin{figure}[!t]
\centering
\includegraphics[width=2.9in]{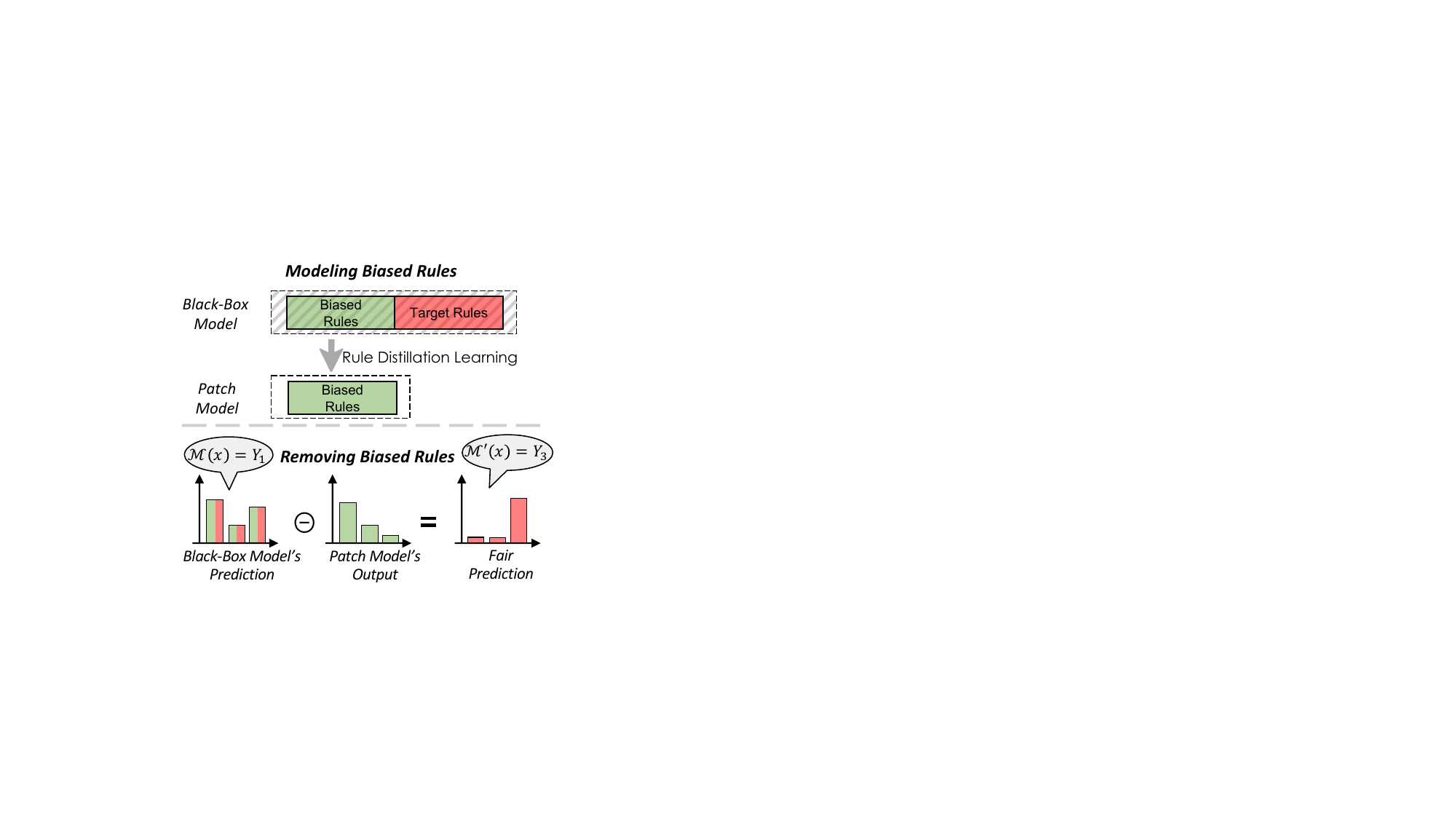}
\caption{Illustration of the proposed Eraser. In the Distill stage, biased rules are distilled and imparted to the patch model. In the Remove stage, the bias response extracted by the patch model is removed from the model output.}
\label{fig_intro}
\vspace{-10pt}
\end{figure}

Although the theoretical foundation of Eraser has indeed provided a direction for debiasing deployed models, obtaining  $p(y|b)$, the response of biased rules in the model, at the inference stage is challenging. This is primarily due to our interaction with the model being limited to probing its output, $p(y|\mathbf{x})$, which is a mixed result of both target and biased rules, rather than a pure biased rule. To address this issue, we propose a specific implementation of the proposed Eraser, which involves two stages: Distill and Remove, as illustrated in Fig.~\ref{fig_intro}. The Distill stage is essentially a preparatory phase for obtaining biased rules, while the Remove stage involves the direct application of the remove strategy from Eraser during the inference. In the \emph{Distill} stage, we introduce rule distillation learning that distills biased rules from a mix of multiple rules in the deployed model and transfers the distilled biased rules to an additional patch model. Specifically, we propose a causality-based distillation strategy, utilizing sample pairs extracted from a small calibration set to distill the biased rules from the deployed model (refer to Sec.~\ref{Distillation} for details).
The underlying mechanism is to eliminate the confounding effect introduced by target rules. Since distilling $p(y|b)$ requires supervised data, it is not feasible to directly distill $p(y|b)$ from test samples. Therefore, we utilize an additional small model, referred to as the patch model, to learn the distilled biased rules. In the subsequent \emph{Remove} stage, the patch model is capable of extracting $p(y|b)$ for test samples. We then subtract the logarithmic value of $p(y|b)$ from the model's logit output to obtain the final unbiased result. The collaboration between the two stages, facilitated by the patch model as an intermediary, enables the debiasing of deployed models.

The Inference-Time Rule Eraser offers a practical solution for debiasing deployed models by removing biased rules from model output. Extensive experimental evaluations demonstrate that Eraser has superior debiasing performance for deployed models, even outperforming those methods that incorporate fairness constraints into the training algorithm for retraining. Furthermore, While Eraser's primary focus is on fairness issues, its effectiveness in addressing general bias concerns has also been validated. This highlights the adaptability and practicality of Eraser in addressing problems instigated by spurious prediction rules.

The contributions of this paper are summarized as follows:
\begin{itemize}

    \item We theoretically derived the Inference-Time Rule Eraser (Eraser), a method that removes biased rules and retains target rules by solely adjusting the model output, without the need for access to model parameters.
    
    \item We further propose \textit{Rule Distillation Learning}, which employs a small number of samples to distill biased rules from a black-box model. These distilled biased rules are stored in a patch model to capture the responses of biased rules about test samples. 

    \item To the best of our knowledge, we are the first in the field of machine learning, not just limited to fairness, to explicitly edit the predictive rules from the model output without modifying the model parameters.

\end{itemize}

\section{Model Debiasing}
\subsection{Fairness in Machine Learning} 
The problem of model bias in deep learning has 
received increasing attention in recent years~\cite{DBLP:journals/tmm/OuyangHLCLLL22,DBLP:journals/tmm/WenNLWTW24,DBLP:journals/tmm/YangZYD23,DBLP:journals/tmm/SongYWX24,DBLP:journals/tmm/YaoMZX22,DBLP:journals/tmm/HanSDWLN23,DBLP:journals/tmm/LiuSSYHS24}.
Methods for measuring bias, such as Equalodds~\cite{hardt2016equality}, have revealed that AI models can exhibit societal bias towards specific demographic groups, including gender and race~\cite{yao2017beyond,zhao2017men}. Strategies to promote fairness can be broadly classified into pre-processing, in-processing, and post-processing methods~\cite{caton2020fairness}, which correspond to interventions on the training data, the training algorithm, and the trained model, respectively. Among them, in-processing has received the most research attention and has demonstrated the most effective debiasing performance, while post-processing methods have been the least studied.

Pre-processing methods are designed to modify the distribution of training data, making it difficult for the training algorithm to establish a statistical correlation between bias attributes and tasks~\cite{kamiran2012data,ramaswamy2021fair}. For example, study~\cite{kamiran2012data} suggests adjusting the distribution of training data to construct unbiased training data. Another study~\cite{ramaswamy2021fair} proposes the generation of new samples to modify the distribution of training data through data augmentation. These strategies aim to decouple the target variable in the training data from the bias variable, thereby preventing the learning of bias in the training of the target task. In-processing methods focus on incorporating additional fairness-aware constraints during ML training~\cite{wang2019balanced,park2022fair,tartaglione2021end}. For instance, research~\cite{wang2019balanced} introduces adversarial fairness penalty terms, which minimize the learning of features related to bias variables during the learning process of the target task, thereby mitigating bias. Some studies~\cite{park2022fair,tartaglione2021end} propose fairness-aware contrastive learning loss that brings samples with different protected attributes closer in feature space, thereby reducing differential treatment by the model. Despite their effectiveness, these methods generally do not extend to debiasing deployed models without retraining.

The field of post-processing is primarily concerned with the calibration of pre-trained machine learning models to ensure fairness~\cite{wang2022fairness,zhang2022fairness}. A common approach in this area involves adjusting the predictions of models to align with established fairness benchmarks\cite{hardt2016equality,pleiss2017fairness}. For instance, some methods modify model outputs directly to satisfy the Equalodds standard by solving an optimization problem~\cite{pleiss2017fairness}. The study by Hardt et al. proposes the learning of distinct classification thresholds for various groups or alterations in output labels to fulfill fairness criteria~\cite{hardt2016equality}. However, these techniques require explicit bias labels during testing as the thresholds and output transformation probabilities are group-specific. An alternative approach~\cite{kim2019multiaccuracy} approximates the impact of fine-tuning models through optimizing parameters added to the model. A study~\cite{sun2022causality} suggests modifying the weight of neurons, identified by causality-based fault localization, to eliminate model bias. Another study~\cite{KirichenkoIW2023} employs reweighted features to fine-tune the last layer of parameters, thereby reducing reliance on biased features. However, these methods necessitate modifying model parameters. In addition, ~\cite{lohia2019bias} suggests adjusting model predictions by identifying bias in model outputs and modifying protected attributes accordingly, which aids in achieving individual and group fairness standards on tabular data. Nevertheless, this method requires altering protected attributes of test examples during testing, which poses challenges for computer vision data. Recent research introduces a pixel perturbation network that adds extra pixel perturbations to the input image to minimize bias~\cite{wang2022fairness,zhang2022fairness}. However, this method requires additional access to the model’s parameters to perform gradient backpropagation for training the perturbation network.

Unlike existing methods, we investigate a debiasing approach that relies solely on model output, without requiring access to model parameters or any labels of test samples. 

\subsection{Rule Editing}
Rule Editing~\cite{santurkar2021editing} provides a lightweight approach to modifying the rules or knowledge embedded in pre-trained models, bypassing the need for full retraining. 

Model fine-tuning is a straightforward approach to rule editing. It involves using data that complies with the desired rules to fine-tune the model. For instance, some studies~\cite{DeCao2021EditingFK} fine-tune pre-trained language models to correct inaccuracies in their knowledge, significantly reducing the computational costs compared to retraining. Other research~\cite{sun2022singular} suggests further reducing costs by adjusting eigenvalues after performing singular value decomposition (SVD) on the backbone network. Research~\cite{santurkar2021editing} proposes generating counterfactuals for erroneous rules in training data and using them for fine-tuning to correct the model's rules.
Local parameter modification~\cite{sun2022causality,singla2022salient} focuses on adjusting specific model parameters related to the rules to be edited. 
In addition to fine-tuning the identified local parameters, other research~\cite{zhao2021ai} proposes directly reversing the neural activation encoded by the located parameters without training.
Structural expansion is another approach to address model editing. Research~\cite{fu2022sound} leverages the properties of ReLU and proposes a theoretically guaranteed repair technique.

However, these methods require adjustments to the structure or parameter weights of white-box models, posing challenges for editing rules in black-box scenarios. 

\section{Inference-Time Rule Eraser}

In this section, we take a close look at the biased deployed model from the perspective of Bayesian inference. We then introduce Inference-Time Rule Eraser, a flexible method conceived from an output probabilistic standpoint, aimed at removing biased rules from the model’s output without necessitating any alterations to the model parameters.

\subsection{Problem Formulation}
The problem of debiasing deployed models can be formally defined as follows: Given an already-trained deployed model, denoted as $\mathcal{M}$, which has been trained on a large dataset $\mathcal{D}_{L}$. This model $\mathcal{M}$ serves as a black-box machine learning service that takes an input $\mathbf{x} \in X$ and produces a probabilistic output $\mathcal{M}(\mathbf{x})$, where $\mathcal{M}(\mathbf{x})^j$ represents the probability of predicting $\mathbf{x}$ as the $j^{th}$ target class. The training dataset $\mathcal{D}_{L}$ and the model parameters are inaccessible. To perform debiasing on $\mathcal{M}$, we are provided with a small-scale dataset called the \textit{calibration set} $\mathcal{D}_{S}=\{\mathbf{x}_i,y_i,b_i\}_{i=1:N}, |\mathcal{D}_{S}|\ll|\mathcal{D}_{L}|$. Here $\mathbf{x}_i\in X_{S}$ denotes the $i^{th}$ input feature, $y_i\in Y$ denotes its target label, $b_i\in B$ denotes its bias label (e.g., gender), and $|\mathcal{D}_{S}|\ll|\mathcal{D}_{L}|$ represents the size of $\mathcal{D}_{S}$ is significantly smaller than the training set $\mathcal{D}_{L}$ of the deployed model. 
Subject to the constraint of not altering the parameters of the deployed model $\mathcal{M}$, the objective of debiasing is to ensure that the model's output remains independent of bias-related information in the input, while preserving accuracy in predicting the target label.

\subsection{Bayesian Inference}
The issue of fairness fundamentally involves two variables: the target variable $Y$ and the bias variable $B$. Both variables shape the generation of data $X$.
For instance, image data $\mathbf{x}$ may contain both the target feature $\mathbf{x}^y$ (i.e., the feature representing the target variable $y$) and the bias feature $\mathbf{x}^b$ (i.e., the feature representing the bias variable $b$, such as male or female). 

Fair machine learning models should base predictions of $y$ exclusively on the target feature $\mathbf{x}^y$ of the input $\mathbf{x}$. However, the inherent nature of machine learning is to learn all features that are beneficial for optimizing the likelihood $p(y|\mathbf{x})$. In the biased training dataset that has a high correlation between the target variable and the bias variable, i.e., the prior $p(y|b)$ is imbalanced, machine learning models often tend to make decisions based on bias features $\mathbf{x}^b$ of the input $\mathbf{x}$, which causes the model to be \emph{biased}.

\myparagraph{Bayesian Inference of Biased Models.}
\hspace{2mm}
In machine learning, we are generally interested in estimating the conditional probability $p(y|\mathbf{x})$. From a Bayesian perspective, the conditional probability $p(y|\mathbf{x})$ of biased models can be interpreted as:
\begin{equation}\label{bayes1}
     p(y=j|\mathbf{x})=p(y=j|\mathbf{x},b) =\frac{p(\mathbf{x}|y=j,b)}{p(\mathbf{x}|b)} p(y=j|b),
\end{equation}
where the conditional probability $p(y|\mathbf{x})$ can be expressed as $p(y|\mathbf{x},b)$. This is because, although $b$ is not directly fed into the model, it can be regarded as an implicit input due to its impact on the generation process of $\mathbf{x}$. The probabilities $p(y|b)$ and $p(\mathbf{x}|b)$ represent the unknown data distribution of the biased model’s training set $\mathcal{D}_{L}$.

\myparagraph{Bayesian Inference of Fair Models.}
\hspace{2mm}
Fair machine learning models, on the other hand, aim to rely solely on target features $\mathbf{x}^y$ strive to optimize the model towards a fair data distribution, denoted as $\hat{P}(X, Y, B)$. This distribution maintains a uniform correlation between $y$ and $b$, i.e., $\hat{p}(y=j|b) = 1/k$, where $k$ is the number of classes in the target variable. This uniform correlation indicates that there is no statistical association between $y$ and $b$.
For $\hat{P}(X, Y, B)$, the conditional probability of $y$ given $\mathbf{x}$, denoted as $\hat{p}(y=j|\mathbf{x})$, can be decomposed using Bayesian interpretation:
\begin{equation}\label{bayes2}
\begin{aligned}
     &\hat{p}(y=j|\mathbf{x})=\hat{p}(y=j|\mathbf{x},b)=\frac{\hat{p}(\mathbf{x}|y=j,b)}{\hat{p}(\mathbf{x}|b)} \hat{p}(y=j|b),\\ &\hat{p}(y=j|b)= 1/k.
\end{aligned}
\end{equation}
The reason $\hat{p}(y=j|b) = 1/k$ is because fair machine learning aims to optimize towards a data distribution where the target variable $y$ and bias variable $b$ are independent of each other. These modeling requirements of fair machine learning can be observed in fairness metrics such as Equalodds~\cite{hardt2016equality}.

Assuming that all instances in the training dataset of the biased model and the fair data distribution are generated from the same process $p(\mathbf{x}|y, b)$, i.e., $p(\mathbf{x}|y, b) = \hat{p}(\mathbf{x}|y,b)$, there can still be a discrepancy between the training set of biased models and the fair data distribution due to differences in the conditional distribution $p(y|b)$ and evidence $p(\mathbf{x}|b)$. This discrepancy in the two Bayesian interpretations reflects the influence of the biased rule in the biased model. By eliminating the difference in the two Bayesian interpretations while retaining the learning of biased models for $p(\mathbf{x}|y, b)$, we can remove the biased rule from the biased model and retain the target rule. To achieve this, we introduce the Inference-Time Rule Eraser in the following section.

\subsection{Inference-Time Rule Eraser}
In machine learning, the model’s inference for input sample $\mathbf{x}$ is essentially an estimation of the conditional probability $p(y|\mathbf{x})$. This conditional probability can be modeled as a multinomial distribution $\phi$:
\begin{equation}
      \phi=\phi_1^{\mathbf{1}\{y=1\}} \phi_2^{\mathbf{1}\{y=2\}} \cdots \phi_k^{\mathbf{1}\{y=k\}} ; \quad \phi_j=\frac{e^{\eta_j}}{\sum_{i=1}^{k} e^{\eta_i}},
\end{equation}
where $\mathbf{1}\{\cdot\}$ denotes the indicator function.
The Softmax function maps the model’s class-$j$ logits output, denoted as $\eta_j$, to the conditional probability $\phi_j$. 

From a Bayesian inference perspective, $\phi_j$ can be interpreted according to the Bayes theorem as presented in Eq.~\eqref{bayes1}. To distinguish between the outputs of biased models and fair models, we use $\eta_j$ and $\phi_j$ to represent the logit for class-$j$ and conditional probability $p(y=j|\mathbf{x})$ for biased models' outputs, respectively. Similarly, we use $\hat{\eta}_j$ and $\hat{\phi}_j$ to represent the logit for class-$j$ and conditional probability $\hat{p}(y=j|\mathbf{x})$ for fair models' outputs, respectively.

To remove biased rule and retain only the target rule in the outputs, we introduce the Inference-Time Rule Eraser:

\begin{theorem} (Inference-Time Rule Eraser)
Assume $\hat{\phi}$ to be the conditional probability of the fair models that without biased rule, with the form $\hat{\phi}_j=\hat{p}(y=j|\mathbf{x})=\frac{p(\mathbf{x}|y=j,b)}{p(\mathbf{x}|b)} \frac{1}{k}$, and $ \phi $ to be the conditional probability of the biased (deployed) model that with biased rule, with the form $ \phi_j=p(y=j|\mathbf{x})=\frac{p(\mathbf{x}|y=j,b)}{p(\mathbf{x}|b)} p(y=j|b)$. If $\phi$ is expressed by the standard Softmax function of model output $\eta$, then $\hat{\phi}$ can be expressed as
\begin{equation}\label{theorem1}
     \hat{\phi_j} = \frac{e^{\log(\phi_j)- \log(p(y=j|b))} } {\sum_{i =1}^k e^{\log(\phi_i) - \log( p(y=i|b))} } 
\end{equation}
\end{theorem}

\noindent{\bf Proof.} The conditional probability of a categorical distribution can be parameterized as an exponential family. It gives us a standard Softmax function as an \emph{inverse parameter mapping}:
\begin{equation}\label{pr1}
      \quad \phi_j=\frac{e^{\eta_j}}{\sum_{i=1}^k e^{\eta_i}},  \quad \hat{\phi_j}=\frac{e^{\hat{\eta_j}}}{\sum_{i=1}^k e^{\hat{\eta_i}}}
\end{equation}
and also the \emph{canonical link function}:
\begin{equation}\label{pr2}
\eta_j=\log \frac{\phi_j}{\phi_k}, \hat{\eta}_j=\log \frac{\hat{\phi}_j}{\hat{\phi}_k}
\end{equation}

\noindent Given that all instances in the data distribution of biased models and fair models are generated from the same process $p(\mathbf{x}|y,b)$, i.e., $p(\mathbf{x}|y,b) = \hat{p}(\mathbf{x}|y,b)$. Thus we can combine Eq.~\eqref{bayes1} and Eq.~\eqref{bayes2}:
\begin{equation}\label{pr3}
\hat{\phi}_j=\phi_j \cdot \frac{\hat{p}(y=j|b)}{p(y=j|b)} \cdot \frac{p(\mathbf{x}|b)}{\hat{p}(\mathbf{x}|b)}
\end{equation}
where $\hat{p}(y=j|b) = 1/k$.

\noindent Subsequently, we apply Eq.~\eqref{pr3} to Eq.~\eqref{pr2}:
\begin{equation} \label{equ0507}
    \hat{\eta}_j=\log \frac{\phi_j \cdot p(\mathbf{x}|b)}{\hat{\phi}_k \cdot p(y=j|b) \cdot k \cdot \hat{p}(\mathbf{x}|b)}
\end{equation} 

\noindent Then, combining Eq.~\eqref{equ0507} and Eq.~\eqref{pr1}, we have:
\begin{equation}
    \hat{\phi_j} 
    =\frac{e^{\eta_j - \log p(y=j|b) - \log\frac{k \cdot \hat{p}(\mathbf{x}|b)}{p(\mathbf{x}|b)}}}{\sum_{i=1}^k e^{\eta_i -\log p(y=i|b) - \log\frac{k \cdot  \hat{p}(\mathbf{x}|b)}{p(\mathbf{x}|b)}}} 
    = \frac{e^{\eta_j - \log p(y=j|b) }}{\sum_{i=1}^k e^{\eta_i -\log p(y=i|b)}} 
\end{equation}
and since $\eta_j=\log \frac{\phi_j}{\phi_k}$, we have:
\begin{equation}
     \hat{\phi_j} = \frac{e^{\log(\phi_j)- \log(p(y=j|b))} } {\sum_{i =1}^k e^{\log(\phi_i) - \log(p(y=i|b))} } 
\end{equation}

\begin{figure*}[t!]
\setlength{\abovecaptionskip}{0.6em} 
\centering
\includegraphics[height=185px]{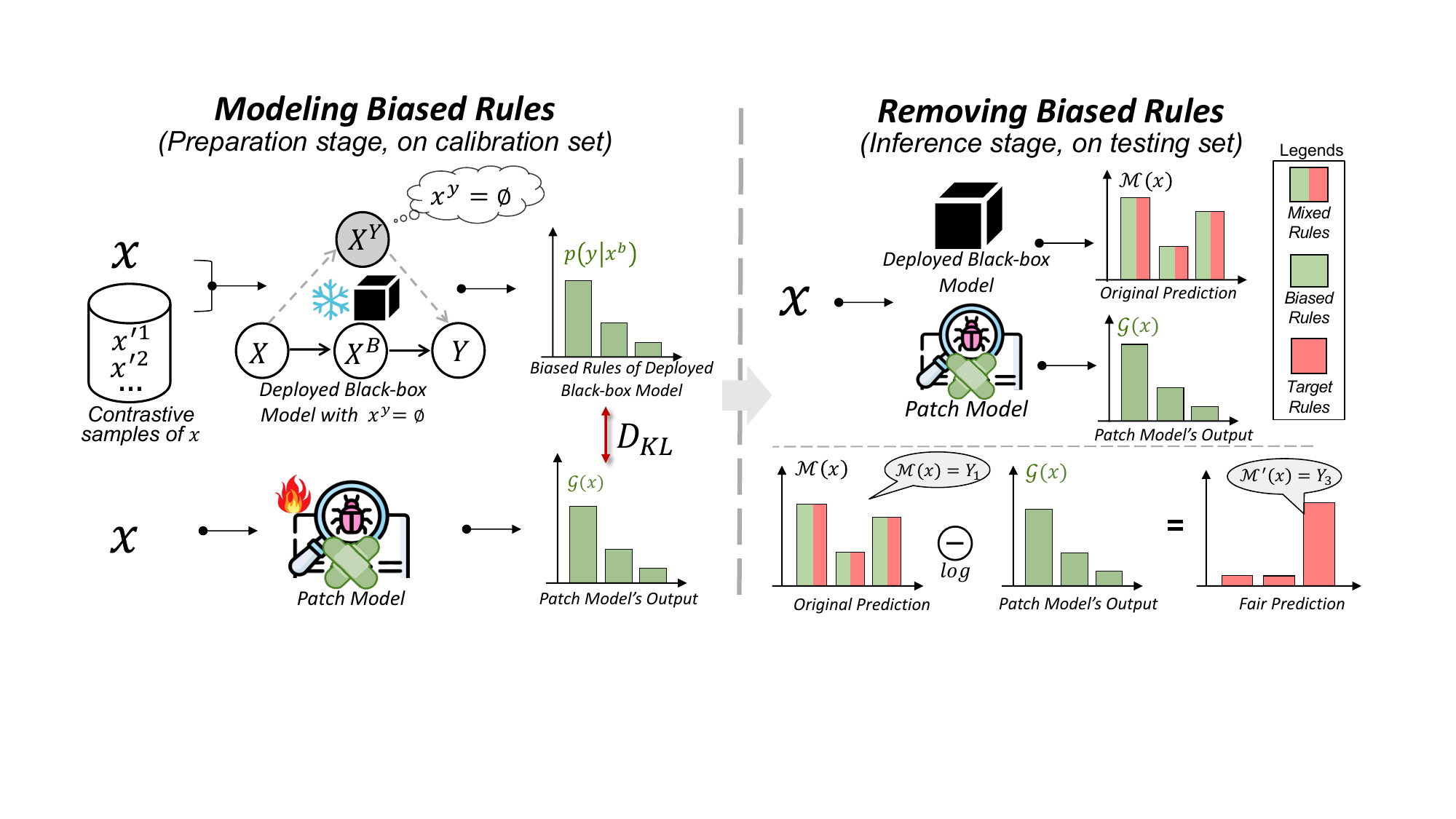}
\caption{The architecture of our proposed Inference-Time Rule Eraser (Eraser). During preparation, the Eraser uses a causality-based distillation strategy to distill biased rules from the deployed black-box model and stores them in the additional patch model. During inference, the Eraser subtracts the patch model's output from the black-box model's original output in \textit{log} space to produce a fair prediction.}
\label{method}
\vspace{-10pt}
\end{figure*}

Theorem 1 shows that debiasing deployed models can be accomplished through adjustments to the model output. 
It proposes that the influence of biased decision rules $p(y|b)$ can be removed by subtracting them in logarithmic ($\log$) space from the model outputs, enabling debiasing without the need to modify the model parameters.
While the Inference-Time Rule Eraser indeed illuminated the path toward debiasing for deployed models, deriving $p(y|b)$ from the unknown training data distribution of the deployed model presents a challenge due to our lack of access to these data distributions. Essentially, the deployed model is a parameterized representation of the conditional probability $p(y|\mathbf{x})$ in training data distribution. Since the bias feature $\mathbf{x}^b$ related to $b$ is part of $\mathbf{x}$, the deployed model inherently models $p(y|b)$. 
In the following section, we will explore how to obtain $p(y|b)$ from the deployed model and apply the Inference-Time Rule Eraser.

\section{The Proposed Implementation}
\subsection{Overview}
The Inference-Time Rule Eraser mitigates bias in deployed models by removing the biased rule $p(y|b)$ from the log space of the model's output. Although the deployed model inherently includes $p(y|b)$, extracting this information directly from the deployed model presents challenges. To address this, we propose \textit{Rule Distillation Learning}, inspired by the concept of Knowledge Distillation. This method comprises the following steps: (1) distilling the biased rule from the deployed model using a small calibration set based on concepts from causal inference, and (2) training a supplementary small model called a patch model to learn these rules. 

The reason for training the patch model to learn the distilled biased rules is that distilling these rules requires labeled samples, which are unavailable during the inference phase for test samples. By training the patch model to learn the distilled biased rules, the patch model can process test samples $\mathbf{x}$ as input and produce output reflecting the biased rule $p(y|b)$ without requiring additional annotations.

Finally, as illustrated in Fig.\ref{method}, our complete bias mitigation method consists of two stages that both utilize the patch model as an intermediary. In the first stage, known as the preparation stage, \textit{Rule Distillation Learning} is used to extract pure biased rules from the deployed model and transfer them to the patch model, as explained in Sec.~\ref{Distillation}. In the second stage, the patch model is capable of extracting $p(y|b)$ for test samples. We then subtract the logarithmic value of $p(y|b)$ from the model's logit output to obtain the final unbiased result. This process is detailed in Sec.\ref{Removing}.

\subsection{Rule Distillation learning}~\label{Distillation}
In the following sections, we will elaborate on Rule Distillation Learning, beginning with a causal analysis of biased rules, followed by a detailed account of how it \textit{distills} the biased rules from the deployed model and guides the additional patch model to \textit{learn} the distilled rules.

\begin{figure}[!t]
\centering
\includegraphics[width=2.4in]{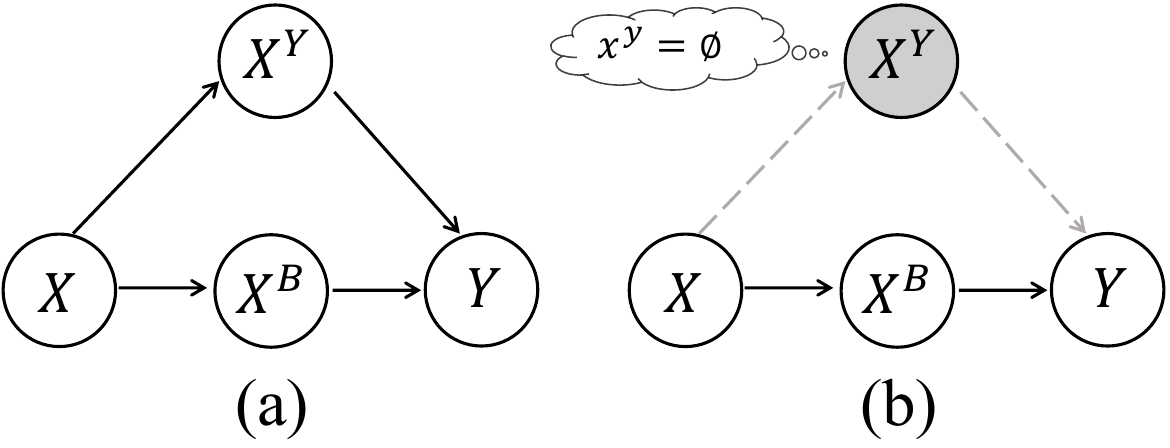}
\caption{The causal graph of the inference process of the biased model. (a) The output $Y$ of the biased model is directly affected by the target feature $X^Y$ and the bias feature $X^B$ in the input $X$. (b) With the conditioning $x^y = \emptyset$, the output $Y$ is only affected by the bias feature $X^B$.}
\label{fig_2causal}
\vspace{-10pt}
\end{figure}

\myparagraph{Causal Perspective.}
\hspace{2mm}
We begin by visiting the causal relationship between the input feature $\mathbf{x} \in X$ and the model output $y \in Y$ within the deployed model. Fig.~\ref{fig_2causal}(a) illustrates the causal graph of the inference process in the deployed model, where the model’s output $y \in Y$ relies on both features $\mathbf{x}^y \in X^Y$ about target label $y \in Y$ and features $\mathbf{x}^b \in X^B$ associated with bias label $b \in B$ in input $\mathbf{x} \in X$. Here, $X^B \rightarrow Y$ represents the contribution of bias information to the target output $p(y|\mathbf{x}^b)$, where $p(y|\mathbf{x}^b)$ is equivalent to $p(y|b)$ \footnote{Below we use the terms $p(y|b)$ and $p(y|\mathbf{x}^b)$ interchangeably.} because the deployed model’s use of bias information $b$ can only be through $\mathbf{x}^b$. 
For subsequent bias removal, our focus is on $p(y|\mathbf{x}^b)$. However, the output generated by providing input $\mathbf{x}$ to the deployed model $\mathcal{M}$ is not $p(y|\mathbf{x}^b)$, but rather a mixed response of $\mathbf{x}^y$ and $\mathbf{x}^b$:
\begin{equation}
\mathcal{M}(X=\mathbf{x})^j= p(y=j|\mathbf{x}^y,\mathbf{x}^b)
\end{equation}
where $\mathcal{M}(X = \mathbf{x})^j$ denotes the model's probabilistic output for the $j^{th}$ target class given the input $\mathbf{x}$.

In causal theory, this situation arises because the path $X \rightarrow X^Y \rightarrow Y$ acts as a backdoor path for the path $X \rightarrow X^B \rightarrow Y$, causing $X^Y$ to become a confounder. Consequently, the causal effect from $X$ to $Y$ becomes a mixture of these two paths rather than focusing solely on the path $X^B \rightarrow Y$. 
To block the path $X \rightarrow X^Y \rightarrow Y$, as depicted in Fig.~\ref{fig_2causal}(b), we utilize the concept of \textit{conditioning} in causal theory~\cite{pearl2016causal}, by actively conditioning on $X^Y$, ensuring accurate assessment of the causal relationship $X^B \rightarrow Y$. Specifically, we set $X^Y$ as an empty set, i.e., $\mathbf{x}^y = \emptyset$, thereby obstructing the confounding influence of $X^Y$ on $Y$. By employing $\mathbf{x}^y=\emptyset$, the model input will exclusively comprise the bias features $\mathbf{x}^b$.

With $\mathbf{x}^y=\emptyset$, the conditional probability $p(y=j|\mathbf{x}^b)$ can be considered as the output of model $\mathcal{M}$ with input $\mathbf{x}^b$:
\begin{equation}\label{casualequ}
p(y=j|\mathbf{x}^b) = \mathcal{M}(X =\mathbf{x}^b)^j
\end{equation}
Here, $\mathcal{M}(X = \mathbf{x}^b)$ denotes the model output when only the bias features $\mathbf{x}^b$ are input to the model $\mathcal{M}$, while the input concerning $\mathbf{x}^y$ is null. As a result, the path $X \rightarrow X^Y \rightarrow Y$ is blocked, and the model output $Y$ is influenced solely by the bias features $X^B$.

\begin{figure}[!t]
\centering
\includegraphics[width=2.9in]{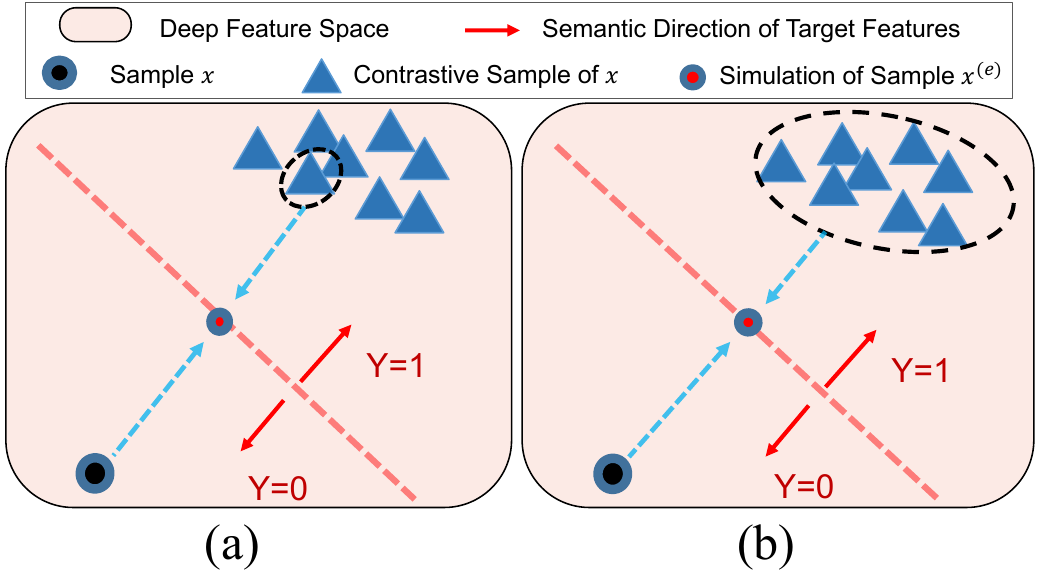}
\caption{Rule distillation via simulating sample-editing. (a) Using a single sample as the contrastive sample of sample $\mathbf{x}$. (b) Employing multiple samples simultaneously as contrastive samples of sample $\mathbf{x}$.}
\label{fig_deepfeature}
\vspace{-10pt}
\end{figure}

\myparagraph{Rule Distillation.}
\hspace{2mm}
To obtain $\mathcal{M}(X=\mathbf{x}^b)$, we need to nullify the $\mathbf{x}^y$ feature in $\mathbf{x}$, resulting in an edited sample $\mathbf{x}^{(e)}$, and then input it into the model $\mathcal{M}$. However, given the difficulty of directly modifying the image $\mathbf{x}$ in pixel space, we adopt an indirect approach.
We utilize $\mathbf{x}$ and contrastive samples to simulate sample editing at the representation layer of model $\mathcal{M}$, ensuring that only $\mathbf{x}^b$ is provided to the model $\mathcal{M}$. This approach is motivated by recent research indicating that features deep in a network tend to be linearized~\cite{upchurch2017deep,bengio2013better}. As shown in Fig.~\ref{fig_deepfeature}(a), using a target class number $k$=2 as an example, there exists a semantic direction for modifying target features in the deep feature space (i.e., the model’s representation space $h(\cdot)$). 
We aim to simulate the result of feeding the edited sample $\mathbf{x}^{(e)}$ into the model $\mathcal{M}$. 
If the deep feature $h(\mathbf{x})$ of the original sample $\mathbf{x}$ moves along the aforementioned semantic direction towards a state away from any target class, this will result in maximum uncertainty regarding target information. The resulting deep feature can then be regarded as the feature representation $h(\mathbf{x}^{(e)})$ of $\mathbf{x}^{(e)}$.

However, directly obtaining the aforementioned semantic direction is challenging. Therefore, we adopt feature interpolation to simulate the movement of $h(\mathbf{x})$ towards $h(\mathbf{x}^{(e)})$. For sample $\mathbf{x}$, we can select a sample $\mathbf{x}'$ from the calibration dataset $\mathcal{D}_{S}$ based on the labels of samples, which shares the same bias label as $\mathbf{x}$ but has a different target label. We then calculate $h(\mathbf{x}^{(e)})$:
\begin{equation}
h(\mathbf{x}^{(e)})= \frac{1}{2}\left(h(\mathbf{x})+ h(\mathbf{x}')\right)
\end{equation}
where $h(\mathbf{x}^{(e)})$ does not lean towards any particular target class, while simultaneously preserving the information about $\mathbf{x}^b$.
Similarly, this can be generalized to situations where the target class number $k>2$: 
\begin{equation}~\label{overlinex}
h(\mathbf{x}^{(e)})= \frac{1}{k}\big(h(\mathbf{x})+ \sum_{\mathbf{x}' \in J(\mathbf{x})}^{} h(\mathbf{x}')\big)
\end{equation}
Here, $J(\mathbf{x})$ represents the set of contrastive samples for $\mathbf{x}$ that are selected from the calibration set $\mathcal{D}_{S}$. The set $J(\mathbf{x})$ contains $k-1$ samples, each sharing the same bias attribute as $\mathbf{x}$ but having a different target label from $\mathbf{x}$. Additionally, the target labels of samples in $J(\mathbf{x})$ are also distinct from each other.

The output $\mathcal{M}(\mathbf{x})$ of the deployed model $\mathcal{M}$ for an input sample $\mathbf{x}$ can be expressed as $\mathcal{M}(\mathbf{x}) = \text{Softmax}(f(h(\mathbf{x})))$, where $f(\cdot)$ is a linear classifier stacked on top of the representation layer. 
Therefore, combining Eq.~\eqref{overlinex} and Eq.~\eqref{casualequ}, $p(y|\mathbf{x}^b)$ is the result of passing $h(\mathbf{x}^{(e)})$ through the linear classifier $f(\cdot)$ stacked on top of it, followed by the softmax mapping:
\begin{equation}\label{constrativeinference}
\begin{aligned}
p(y|\mathbf{x}^b)&= \text{Softmax}\big(f(h(\mathbf{x}^{(e)}))\big)\\
&= \text{Softmax}\big(f(\frac{1}{k}(h(\mathbf{x})+ \sum_{\mathbf{x}' \in J(\mathbf{x})} h(\mathbf{x}'))\big)\\
&= \text{Softmax}\big(\frac{1}{k}(f(h(\mathbf{x}))+ \sum_{\mathbf{x}' \in J(\mathbf{x})} f(h(\mathbf{x}'))\big)\\
\end{aligned}
\end{equation}

Due to the \emph{canonical link function}, we can replace $f(h(\mathbf{x}))$ with $\log(\mathcal{M}(\mathbf{x}))$. Then,
\begin{equation}
\label{contrastive}
p(y|\mathbf{x}^b)=\text{Softmax}\Big(\frac{1}{k} \log \left(\mathcal{M}(\mathbf{x})\right) + \sum_{\mathbf{x}' \in J(\mathbf{x})}^{} \log\left(\mathcal{M}(\mathbf{x}')\right)\Big)
\end{equation}

Hence, we don’t require access to the model parameters. By simply obtaining the deployed model’s output for the sample $\mathbf{x}$ and its contrast sample $\mathbf{x}'$, i.e., $\mathcal{M}(\mathbf{x})$ and $\mathcal{M}(\mathbf{x}')$, Eq.~\eqref{contrastive} can be used to derive $p(y|\mathbf{x}^b)$.

Moreover, including outlier samples or samples with labeling noise in the construction of the set $J(\mathbf{x})$ can lead to errors in the calculation of $p(y|\mathbf{x}^b)$. To address this issue, as illustrated in Fig.~\ref{fig_deepfeature}(b), we use multiple contrastive samples for each target class that we want to compare, rather than depending on just a single sample:
\begin{equation}\label{multicontrastive}
\begin{aligned}
&p(y|\mathbf{x}^b)=\\ 
&\text{Softmax} \big(\frac{1}{k} (\log(\mathcal{M}(\mathbf{x})) +\sum_{i=1, i\neq y_\mathbf{x}}^{k} \mathop{\scalebox{1.5}{\(\mathbb{E}\)}}_{\mathbf{x}' \in S_{i,b_\mathbf{x}}}\log(\mathcal{M}(\mathbf{x}'))\big)\\
\end{aligned}
\end{equation}
where $y_\mathbf{x}$ and $b_\mathbf{x}$ denote the target label and bias label of the sample $\mathbf{x}$, respectively, and $S_{i, b_\mathbf{x}}$ represents the subset of samples with a target label of $i$ and a bias label of $b_\mathbf{x}$. 

As a result, for each sample $\mathbf{x}$ in the calibration set, we obtain the biased rule $p(y|\mathbf{x}^b)$ with respect to sample $\mathbf{x}$ from the deployed model. For each sample $\mathbf{x}$, $p(y|\mathbf{x}^b)$ is a $k$-dimensional vector corresponding to the $k$ target classes.

\myparagraph{Rule Learning.}
\hspace{2mm}
The rule distillation method outlined in the preceding section necessitates the sample label. However, this information is inaccessible for test samples during the testing phase. 
To obtain biased rules about the test sample during the application phase without any label knowledge about the test sample, we propose the use of an additional patch model, which we refer to as the patch mode, $\mathcal{G}$, to store the biased rules $p(y|\mathbf{x}^b)$ that we have distilled from the deployed model. The output dimensions of $\mathcal{G}$ are set to be the same as those of $\mathcal{M}$.

On the calibration set $\mathcal{D}_s$, we minimize the Kullback-Leibler divergence $\mathcal{D}_{KL}$ between the output of the patch model and $p(y|\mathbf{x}^b)$ regarding $\mathbf{x}$, which has been distilled from the deployed model. This process enables the storage of biased rules into the patch model $\mathcal{G}$:
\begin{equation}\label{Rulelearning}
\mathcal{G} = \underset{\mathcal{G}}{\operatorname{argmin}}  
\mathop{\scalebox{1.6}{\(\mathbb{E}\)}}_{x \in \mathcal{D}_s}
  \mathcal{D}_{K L}\left(\mathcal{G}(\mathbf{x}) \, \| \, p\left(y|\mathbf{x}^b\right)\right)
\end{equation}
Here, $p(y|\mathbf{x}^b)$ about the sample $\mathbf{x}$ has been obtained during the rule distillation phase. Through this rule learning process, the trained model $\mathcal{G}$ can directly provide the biased rules for the test sample during the testing phase.

\subsection{Rule Removing}~\label{Removing}
Our ultimate goal is to eliminate the use of biased rules $p(y|\mathbf{x}^b)$ during model inference. As discussed in the previous section, the patch model has learned the biased rules, distilled from the deployed black-box model. Consequently, the patch model’s response to the input sample $\mathbf{x}$, denoted as $\mathcal{G}(\mathbf{x})$, can replace $p(y|\mathbf{x}^b)$ in the Inference-Time Rule Eraser formula (Eq.~\eqref{theorem1}).

Hence, as illustrated in Fig.~\ref{method}, during the testing stage, we can rectify the bias present in the deployed black-box model by subtracting the patch model's output from the black-box model's output in log space:
\begin{equation}
\label{removeequ}
     P_{fair}(y=j|\mathbf{x}) = \frac{e^{\log(\mathcal{M}(\mathbf{x})^j)- \log(\mathcal{G}(\mathbf{x})^j)} } {\sum_{i =1}^k e^{\log(\mathcal{M}(\mathbf{x})^i) - \log(\mathcal{G}(\mathbf{x})^i)} } 
\end{equation}
Here, $P_{fair}(y=j|\mathbf{x})$ is the final debiasing output, adjusted from the original output of the biased deployed model. $\mathcal{M}(\mathbf{x})^i$ represents the probabilistic output of the deployed model for class $i$, and $\mathcal{G}(\mathbf{x})^i$ represents the probabilistic output of the patch model for class $i$.

As elucidated by the above equation, the proposed debiasing method only requires adjusting the model's output.

\section{Experience}
This section presents the experimental results of Eraser. We begin by introducing the setup, which includes the evaluation metrics, seven datasets, baseline methods for comparison, and implementation details. Following this, we present results on existing datasets and our ImageNet-B, which is built based on the industrial-scale dataset, ImageNet~\cite{deng2009imagenet}. While our experiments are primarily focused on vision tasks, we stress that the core concept and methodology are not limited to them and can efficiently debiasing in structured datasets. 
Furthermore, we extensively analyze the robustness of our method across various factors, including the size of the calibration set and model architecture.

\subsection{Evaluation Setup}
\myparagraph{Points of comparison.}
\hspace{2mm}
In experience, we aim to answer two main questions: (1) How does the performance of models, corrected by Eraser, compare with other methods? (2) How does the fairness of these models? 

To answer the first question, we divide the test set into different groups based on target attributes and bias attributes, then assess accuracy for each group, and then report the {\bf \textit{average group accuracy}} and the {\bf \textit{worst group accuracy}}. To answer the second question, we examine whether the model predictions adhere to the fairness criterion {\bf \textit{Equalodds}}, which measures the changes in model predictions when the bias attributes shift. For example, in bignose and non-bignose recognition, EqualOdds measures \emph{gender} bias as follows:
\begin{equation}
\frac{1}{|Y|} \sum_{y \in Y} \left|\operatorname{Acc}_{\text{group}(Y=y,B=b^{0})}-\operatorname{Acc}_{\text{group}(Y=y,B=b^{1})}\right|
\end{equation}
where $Y$ denotes target labels, $B$ denotes bias attributes such as \emph{male} ($b^{0}$) and \emph{female} ($b^{1}$). $\operatorname{Acc}_{\text{group}(Y=y,B=b^{0})}$ represents the accuracy of the group $Y=y,B=b^{0}$ in the test set.

        
\myparagraph{Datasets.}
\hspace{2mm}
We evaluate our method against the baselines across seven different datasets to ensure comprehensive coverage of various debiasing tasks. However, given the lack of datasets related to multi-class problems in industrial-scale data, we have additionally constructed ImageNet-B to test the debiasing performance.

\textbf{\textit{CelebA}}~\cite{liu2015deep} is a large-scale face image recognition dataset comprising 200,000 images, containing 40 attributes for each image, of which \emph{Gender} is the bias attribute that should be prevented from being uesed for prediction. We chose \emph{Bignose}, \emph{Blonde}, and \emph{Attractive} as target attributes for three recognition tasks to verify our \emph{gender} debiasing effect, as they exhibit the highest Pearson correlation with gender.  \textbf{\textit{UTKface}}~\cite{zhang2017age} is a dataset that contains around 20k facial images annotated with ethnicity and gender. It is used to validate ethnicity debiasing performance by constructing a biased training set with gender as the target attribute. We also used multi-class dataset \textbf{\textit{C-MNIST}}~\cite{hong2021unbiased} to evaluate debiasing. C-MNIST is a dataset of handwritten digits (0-9) with different background colors, where each digit has a strong correlation with the background color. Debiasing aims to eliminate the spurious dependency on background color information in the digit recognition task. 
In addition, we are interested in investigating the debiasing effect of our method when multiple biases coexist. The \textbf{\textit{UrbanCar}} dataset~\cite{li2023whac} presents a unique opportunity as it encompasses two bias attributes: background (\textit{BG}) and co-occurring object (\textit{Co-obj}), with the target attribute being the type of car {urban, country}. 
In addition to the aforementioned visual datasets, we also evaluated debiasing techniques on two structured/tabular datasets, namely \textbf{\textit{LSAC}}~\cite{wightman1998lsac} and \textbf{\textit{COMPAS}}~\cite{angwin2022machine}, both of which are extensively utilized in fairness literature. Specifically, we examined the performance of race debiasing in the context of law school admission prediction for LSAC, and recidivism prediction for COMPAS.


\begin{figure}[!t]
\centering
\includegraphics[width=3.0in]{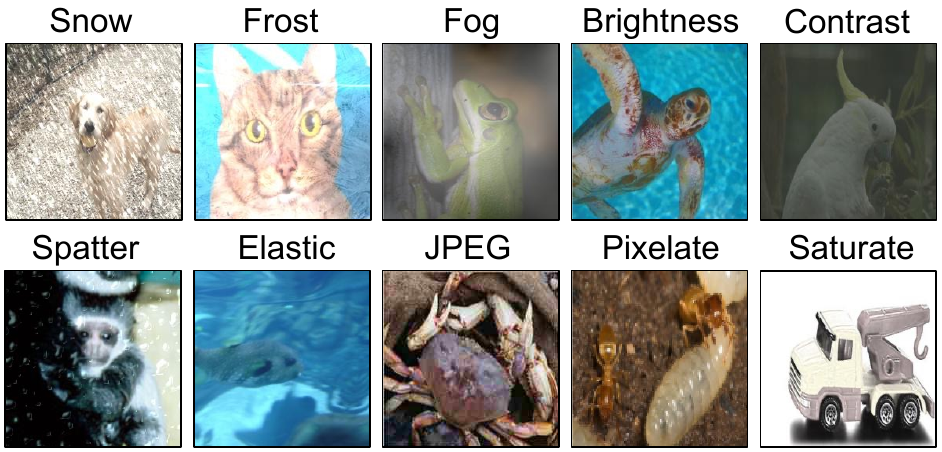}
\caption{Patterns exhibited by the majority (95$\%$) of samples across ten superclasses in the proposed ImageNet-B(ias) dataset.}
\label{imagenetbias}
\vspace{-10pt}
\end{figure}

Since there exists no off-the-shelf industrial-scale dataset specially designed to analyze the debiasing performance in multi-classification tasks, 
we propose \textbf{\textit{ImageNet-B(ias)}}, a dataset aimed at facilitating the study of debiasing in complex tasks. Specifically, we utilize ten types of natural noise patterns~\cite{hendrycks2019robustness} as bias attributes, such as snow, frost, and fog.
We then group semantically similar classes from ImageNet~\cite{deng2009imagenet} into ten super-classes, derived from WordNet~\cite{fellbaum2010wordnet}, to serve as target attributes. In each superclass, 95$\%$ of the images are assigned a specific type of natural noise, while the remaining 5$\%$ of images are randomly assigned one of the other nine types of noise. Consequently, each superclass exhibits a strong correlation with a specific type of natural noise. 
Fig.~\ref{imagenetbias} illustrates the patterns present in the majority of samples in the dataset. For instance, most samples in the dog superclass predominantly exhibit snow noise. 
We utilize the industrial-scale ImageNet-B(ias) dataset to validate our method's debiasing performance in more practical scenarios.

\myparagraph{Baselines.}
\hspace{2mm}
To evaluate the effectiveness of Eraser in debiasing for deployed models, we examined its performance on a range of deployed models with different extents of unfairness. We trained three types of models to serve as the deployed models: the vanilla model, the fair model, and the unfair model. Specifically, (1) The \textit{vanilla} model was trained using cross-entropy loss without the application of any debiasing technique. (2) The fair model was developed by integrating fairness constraints into the model optimization objective. There are two typical methods: \textit{AdvDebias}, which incorporates fairness-related adversarial regularization terms in the loss function~\cite{wang2019balanced}; and \textit{ConDebias}, which leverages contrastive learning to eradicate bias attribute information from the representation~\cite{park2022fair,hong2021unbiased}. (3) The \textit{unfair} model was trained to have the classification ability of both target attributes and bias attributes via multi-task learning~\cite{baxter1997bayesian}, thereby further extracting bias features. 
In addition, there are methods for debiasing after model training.
However, these methods necessitate access to bias labels of test samples or model parameters, requirements not needed by our method. 
We compared our method with these methods under laboratory conditions, i.e., all conditions required by these methods are met, such as model parameters can be modified. These methods include: (1) \textit{Equalodds}:  Solving a linear program to find probabilities with which to change output labels to optimize equalized odds~\cite{hardt2016equality}, (2) \textit{CalEqualodds}: Optimizing over calibrated classifier score outputs to find probabilities with which to change output labels with an equalized odds objective~\cite{pleiss2017fairness}, (3) \textit{FAAP}: Applying perturbations to input images to pollute bias information in the images~\cite{wang2022fairness}, (4) \textit{Repro}: Adding learnable padding to all images to optimize the model output to satisfy fairness metrics~\cite{zhang2022fairness}, (5) \textit{CARE}: Modifying the weight of neurons, identified by causality-based fault localization, to eliminate model bias~\cite{sun2022causality}, (6) \textit{DFR}: Employing reweighted features to fine-tune the last layer of parameters to eliminate the reliance on biased features~\cite{KirichenkoIW2023}. Of these methods, Equalodds and CalEqualodds require known bias labels of test samples. FAAP and Repro need access to model parameters for obtaining gradients. CARE and DFR involve modifying partial model parameters.

\myparagraph{Implementation details.}
\hspace{2mm}
To investigate the debiasing effect on the deployed models, we initially pre-train deployed models. For each target task, we employ four methods (as detailed in the first half of the Baselines section) to train the deployed models, using 5/6 of the training data from each dataset. The default backbone of the models is set to ResNet-34\cite{he2016deep}. To demonstrate the robustness of our method across different architectures, we also use the ViT-B/32~\cite{50650} transformer as the backbone. Subsequently, we use the remaining 1/6 of the training data in each dataset as a calibration set for learning how to debias, such as training the patch model in our method. To maintain a lightweight approach, our patch model uses a smaller backbone, ResNet-18~\cite{he2016deep}.
\input{tables/table1}

\input{tables/table2}
\input{tables/table3}
\subsection{Quantitative Results}
\myparagraph{Main Debiasing performance.}
\hspace{2mm}
Table~\ref{table:1} presents the quantitative results, including accuracy and model bias, of deployed models before and after applying the proposed Eraser method across four datasets.

For the CelebA dataset, we utilize three commonly used target tasks (refer to Tables~\ref{table:big:sub1} to \ref{table:big:sub3}) to evaluate gender debiasing. For the UTKFace dataset, we examine ethnicity debiasing (refer to Table~\ref{table:big:sub4}). In the case of the multi-class datasets, such as C-MNIST and our constructed ImageNet-B, we focus on background color and natural noise debiasing, respectively (refer to Tables~\ref{table:big:sub5} and \ref{table:big:sub6}).

In the CelebA and UTKFace datasets, two types of fair training (AdvDebias and ConDebias) yield models with lower model bias (as measured by Equalodds) compared to the Vanilla model, thanks to the incorporation of fairness constraints. Unfair training, as expected, results in less fair models. For all four model types, the application of our method consistently results in a significant improvement in fairness (reduction in model bias) and an increase in accuracy.
The main observations include: (1) For the \textbf{Vanilla models}, after applying our \textit{Eraser}, the model bias is reduced by more than 70$\%$, reaching up to 89$\%$ at most.
For instance, in the \texttt{Attractive} task, the model bias is reduced from 20.92 to 2.30. 
Alongside the reduction in model bias, average and worst group accuracy also increase, from 76.85 to 78.76 and from 66.20 to 75.73, respectively. 
(2) For the two types of \textbf{fair models}, AdvDebias and ConDebias, our method can further improve the fairness and accuracy of these models. 
For example, for the AdvDebias models in \texttt{Blonde} task, our method still improves fairness (model bias reduction from 11.12 to 5.35) and improves accuracy (from 85.07 to 90.31 and from 61.72 to 85.34 in average and worst group accuracy, respectively). 
This indicates that our method is compatible with the fairness constraints in training. 
(3) For \textbf{unfair model}, Eraser can significantly improve its fairness and accuracy. 
For instance, in the \texttt{BigNose} task, Eraser can decrease model bias to 3.34 while improving average group accuracy to 72.92. 
(4) A comparison of our method versus debiasing during training in Table~\ref{table:1} reveals that a vanilla model integrated with our method can achieve almost superior fairness and accuracy performance than a model that underwent fair training 
(e.g., Vanilla model+Eraser has even better accuracy and model bias than Fair model (AdvDebias and ConDebias)  in \ref{table:big:sub1}). 
This comparison further demonstrates the potential of our method.

The debiasing performance observed in the multi-class datasets (C-MNIST and ImageNet-B), as presented in Tables~\ref{table:big:sub5} and \ref{table:big:sub6}, aligns with the results from the CelebA and UTKFace datasets. After applying our method, four diverse types of deployed models consistently show significant improvements in both accuracy and fairness. 


\myparagraph{Results on structured data.}
\hspace{2mm}
To demonstrate the versatility of our proposed method, we extended our experimentation beyond image data to include structured datasets. 
Table~\ref{table:2} showcases the debiasing performance on two structured datasets, LSAC and COMPAS, both of which are extensively utilized in fairness studies. 
Our method's performance on structured datasets aligns with its performance on image datasets.
By integrating our Eraser with the vanilla model, we consistently achieved a marked increase in fairness and accuracy on both datasets.
For instance, in the LSAC dataset, model bias was reduced significantly from 26.48 to 4.11. Additionally, average accuracy increased from 64.44 to 69.01, while worst group accuracy improved from 19.02 to 57.04.
Furthermore, combining our method with the standard model consistently outperformed standalone fair models, emphasizing the notable debiasing capability of our method. These results on structured datasets suggest that our approach is not limited to specific data types or patterns; it is broadly applicable to both images and structured data.

\myparagraph{Debiasing with multiple biases.}
\hspace{2mm}
In machine learning, it is not uncommon for multiple biases to be present in the training data of models. 
These biases often find their way into the model, leading to biased predictions. 
We conducted experiments to evaluate the effectiveness of our method in simultaneously eliminating multiple biases in a single deployed model.
We conducted experiments on the Urbancars dataset, which concurrently exhibits both background (BG) bias and co-occurring object (Co-obj) bias. 
Initially, we pre-trained the deployed model on the Urbancars dataset using four different training methods. 
Subsequently, for each deployed model, our method trains two patch models, $\mathcal{G}_\text{BG}$ and $\mathcal{G}_\text{Co-obj}$, to counteract the BG and Co-obj biases in the deployed model, respectively.
To mitigate multiple biases, we merge the two patch models to debias simultaneously, referred to as \textit{Eraser}$_\text{\,BG + Co-obj}$. The final debiased output can be formalized as:
\begin{equation}
     P_{fair}(y=j|\mathbf{x}) = \frac{e^{\log(\mathcal{M}(\mathbf{x})_j)- \log(\mathcal{G}_\text{BG}(\mathbf{x})_j)- \log(\mathcal{G}_\text{Co-obj}(\mathbf{x})_j)}} {\sum_{i =1}^k e^{\log(\mathcal{M}(\mathbf{x})_i)- \log(\mathcal{G}_\text{BG}(\mathbf{x})_i)- \log(\mathcal{G}_\text{Co-obj}(\mathbf{x})_i)} } 
\end{equation}
where the symbols have the same meaning as in Eq.~\eqref{removeequ}.

Table~\ref{table:urbancar} shows that using \textit{Eraser}$_{\text{BG + Co-obj}}$ can significantly reduce both types of bias simultaneously. For instance, with the vanilla model, BG bias was reduced from 32.07 to 4.84, while Co-obj bias was reduced from 28.01 to 3.25.
This demonstrates that our method can effectively stack to reduce multiple biases, making it well-suited for complex scenarios involving various biases.

\input{tables/table4}

\subsection{Qualitative Evaluation}

\begin{figure}[t]
\centering
\includegraphics[width=2.8in]{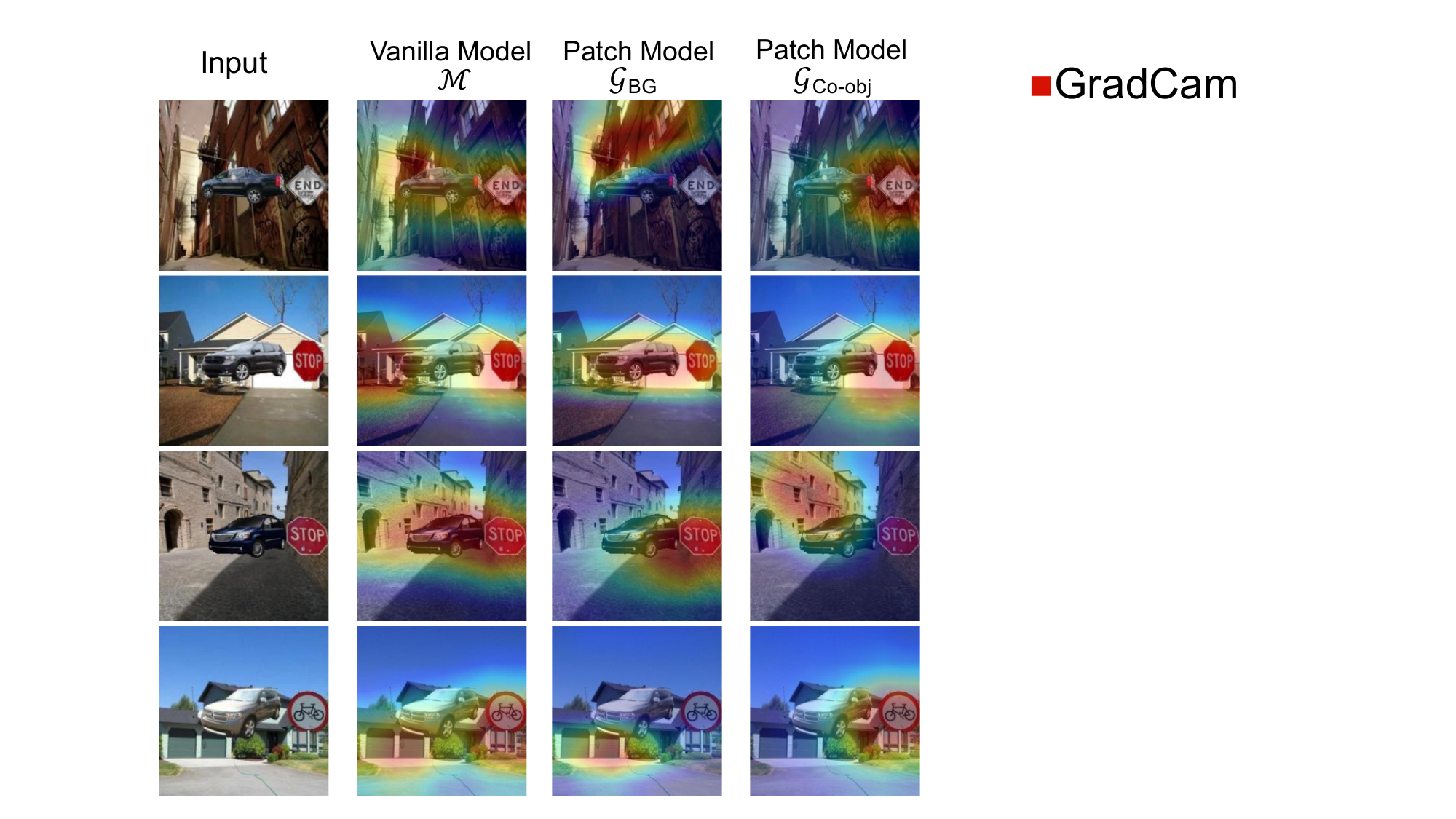}\caption{Grad-CAM results of the Vanilla model and two patch models on UrbanCars dataset.}
\label{gradcam}
\end{figure}

In our quest to provide a comprehensive understanding of our method, we incorporate the model explanation approach \textbf{Grad-CAM}~\cite{selvaraju2017grad} to illustrate our learned patch model successfully captures the bias present in the deployed model. 
For a detailed analysis, we chose the UrbanCars dataset, which is known to contain both types of biases. In Fig.~\ref{gradcam}, we present attention maps of the vanilla (deployed) model and the two patch models designed to capture background (BG) bias and co-occurring object (Co-obj) bias, respectively. The attention maps reveal that the deployed model's focus extends beyond the areas containing cars to include areas with BG and Co-obj elements (e.g., traffic signs). The patch models we trained focus on the BG and Co-obj areas in the images, as expected. This demonstrates that the patch model accurately replicates the biases in the deployed model, confirming that our method can efficiently remove bias without compromising accuracy.

\begin{figure}[t!]
\setlength{\abovecaptionskip}{0.6em} 
\centering
\includegraphics[height=83px]{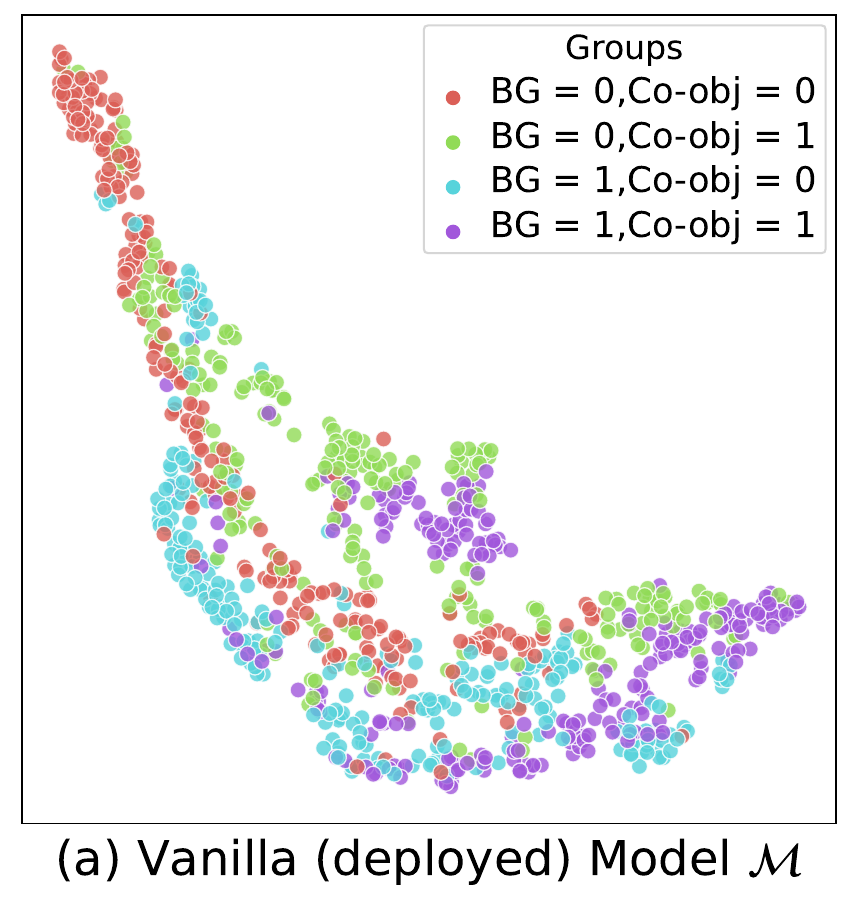}
\includegraphics[height=83px]{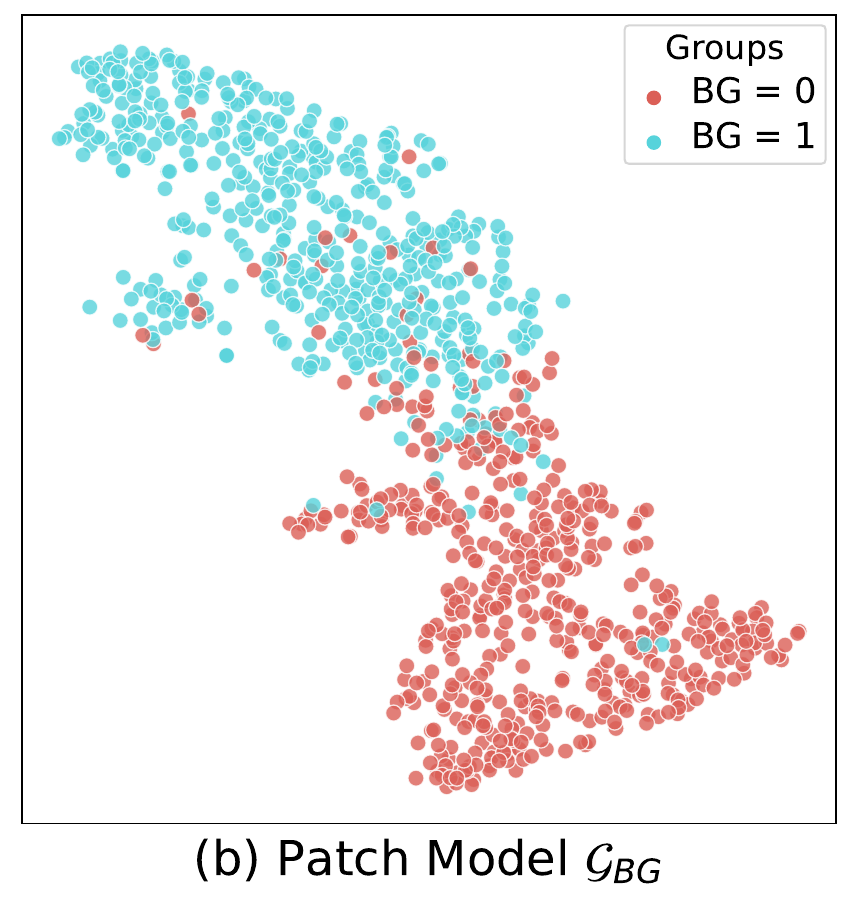}
\includegraphics[height=83px]{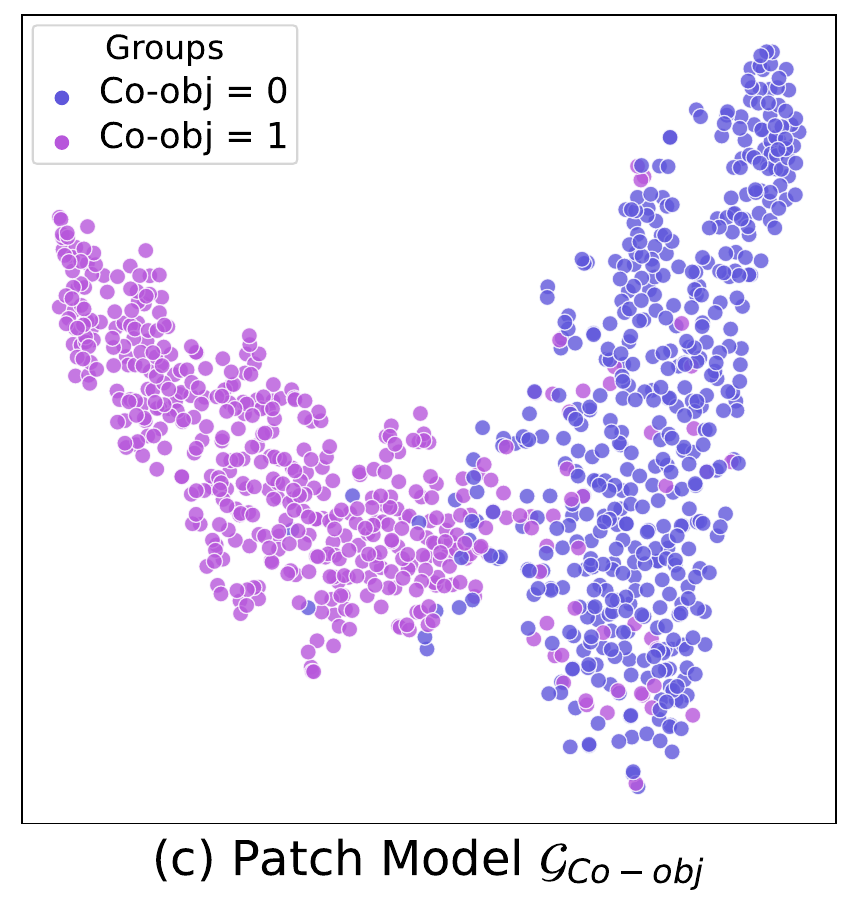}
\caption{T-SNE visualizations of the Vanilla model and two patch models on UrbanCars dataset.}
\label{fig-tsne}
\vspace{-10pt}
\end{figure}

We also present \textbf{t-SNE} visualization of the vanilla model and two patch models on the UrbanCars dataset in Fig.~\ref{fig-tsne}. The points in the figure are divided into different groups based on BG and Co-obj biases, represented by different colors.
In the vanilla model, the representation shows separability for the bias attributes (combinations of the two types of bias, represented by different colored points), indicating that the model relies on biased features for decision-making. Correspondingly, the patch model $\mathcal{G}_\text{BG}$ shows separability for the BG bias, while the patch model $\mathcal{G}_\text{Co-obj}$ exhibits separability for the Co-obj bias. This suggests that the two learned patch models effectively capture the reliance of the vanilla model on these two biased features.

\subsection{Comparison Under Laboratory Conditions}
There exist additional post-training debiasing techniques. However, these methods necessitate access to bias labels of test samples or model parameters, requirements not needed by our method.

To offer a more comprehensive evaluation of our method, we compared our method with these methods in a controlled laboratory setting, i.e., all conditions required by these methods are met, such as model parameters can be accessed. 
Table~\ref{laboratory} presents the quantitative results on image data, where Vanilla represents the deployed model requiring debiasing.
Our proposed method exhibits outstanding performance in terms of both fairness and accuracy across all six tasks.
Take the \texttt{BigNose} recognition task as an example. The vanilla models exhibit substantial fairness issues (model bias = 23.26), with the average group accuracy being a mere 69.48. 
However, our proposed Eraser method significantly reduces the model bias to 3.72 and enhances the worst group accuracy to 75.14.  
Different debiasing techniques show varying degrees of effectiveness in bias mitigation. The Equalodds method slightly outperforms our method in terms of fairness (model bias of Equalodds is 1.50 in the \texttt{BigNose} task) but significantly falls behind in average group accuracy (64.95 for Equalodds versus 75.14 for our method). Additionally, the Equalodds model is only applicable to binary classification tasks and is unsuitable for multi-classification scenarios.
The debiasing capabilities of CARE and DFR are task-dependent. They show limited debiasing performance in the \texttt{BigNose} and \texttt{Attractive} tasks. This may be due to the target features and bias features in the deployed model being more intertwined, making debiasing by modifying some parameters challenging.
In summary, while our method eliminates the need for additional laboratory conditions, it still achieves state-of-the-art performance when compared to other methods.

\subsection{Analysis and discussion}
\myparagraph{Applicability with Limited Calibration Data.} \hspace{2mm}
We explored the relationship between the amount of calibration data used to train the patch model and its impact on the debiasing effect. By adjusting the volume of the calibration set (i.e., the training data for the patch model), we varied the size ratio between the calibration set used for training the patch model and the training set of the deployed model, within a range from 1/1000 to 1/5. A smaller size ratio implies a smaller amount of calibration data for training the patch model. For example, a size ratio of 1/1000 results in a calibration data volume of just 160 images. Importantly, the calibration set used for training the patch model and the training set of the deployed model do not contain overlapping data.

\begin{figure}[t!]
\setlength{\abovecaptionskip}{0.8em} 
\setlength{\belowcaptionskip}{-0.3em} 
\centering
\includegraphics[height=97px]{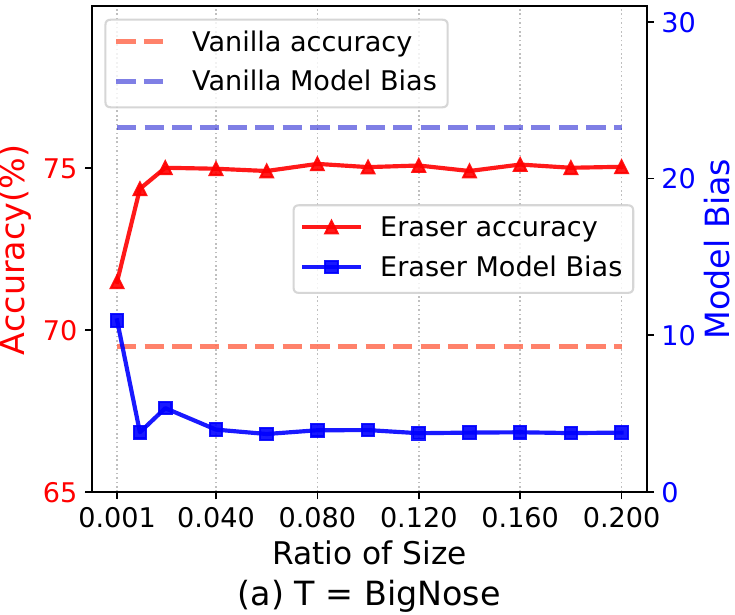}
\includegraphics[height=97px]{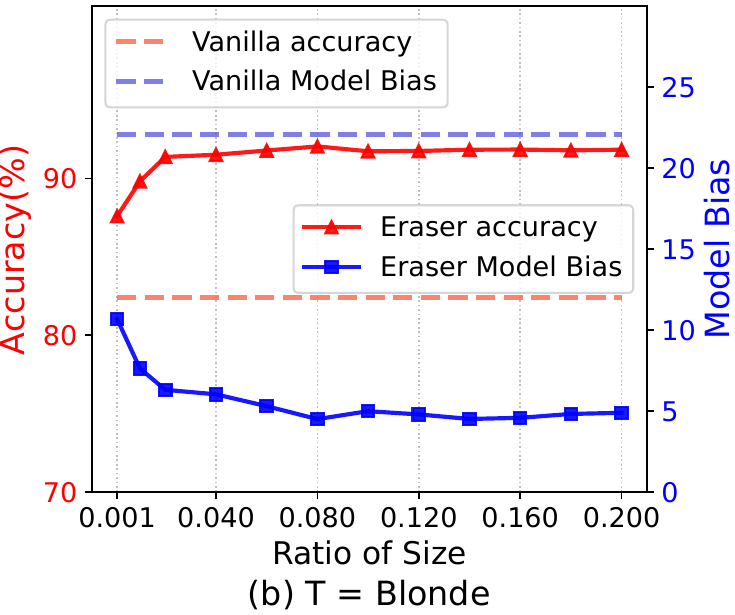}
\caption{Debiasing performance when using different sizes of calibration data.}
\label{fig-sizeratio}
\end{figure}

Fig.~\ref{fig-sizeratio} displays the debiasing results at different size ratios. Our method exhibits significant debiasing performance at different size ratios. For example,  even with a size ratio of 1/1000, we observed over a 50$\%$ reduction in bias.
Moreover, increasing the amount of data used to train the patch model (i.e., a larger size ratio) generally leads to better debiasing performance. However, this trend is not pronounced. 
Taking the target attribute \texttt{BigNose} as an example, the model bias differs by only 7.2, and the accuracy by just 3.4, between size ratios of 1/1000 and 1/5.
These findings suggest that our proposed Eraser method is not highly sensitive to the quantity of calibration data. Even with a small data set (approximately 160 images), our approach remains effective, showcasing its practical applicability.

\begin{figure}[t!]
\setlength{\abovecaptionskip}{0.6em} 
\setlength{\belowcaptionskip}{-0.3em} 
\centering
\includegraphics[height=97px]{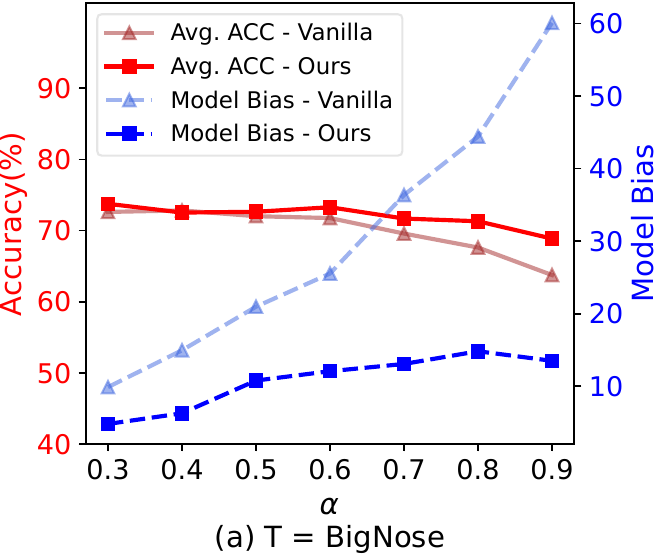}
\includegraphics[height=97px]{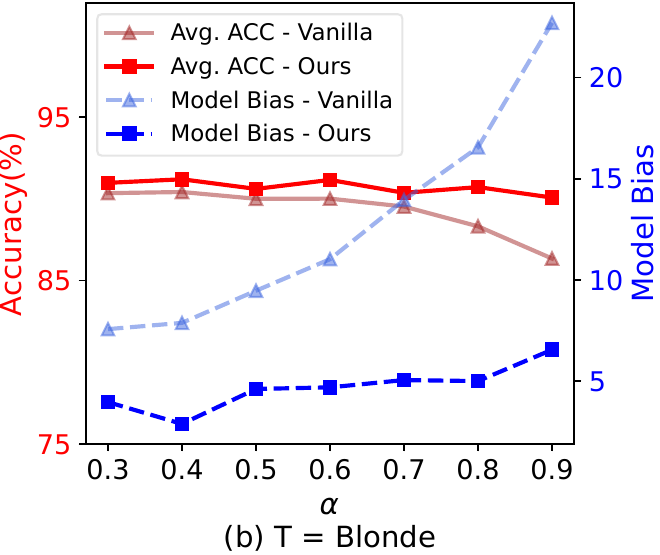}
\caption{Debiasing performance for deployed models with varying levels of model bias, where different deployed models are trained with data of different bias levels; a larger $\alpha$ indicates a higher degree of data bias.}
\label{fig-skew}
\end{figure}

\begin{figure}[!t]
\centering
\includegraphics[width=0.43\textwidth]{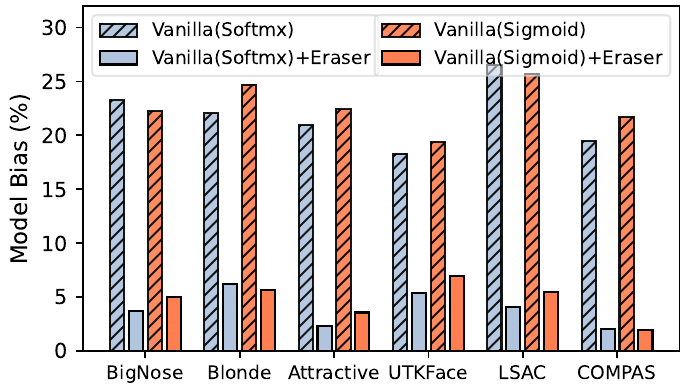}
\caption{Debiasing performance of deployed models using different probability evaluation functions: softmax and sigmoid.}
\label{fig-softmax}
\vspace{-10pt}
\end{figure}

\myparagraph{Robustness to Bias Severity in Deployed Models}\hspace{2mm}
In real-world scenarios, varying degrees of data imbalance in training datasets can lead to differing levels of bias in deployed models.
To assess the effectiveness and robustness of our proposed method in debiasing deployed models derived from different levels of data bias, we created datasets with varying degrees of data imbalance using the CelebA dataset. 
The level of data imbalance, controlled by the $\alpha$, signifies the ratio between the number of minority class samples and majority class samples (with biased labels distinct from the minority class), where a smaller $\alpha$ indicates a higher degree of data imbalance.

Next, we trained different deployed models using these datasets with varying levels of $\alpha$ and evaluated the effectiveness of our approach in mitigating biases in these deployed models.
Fig.~\ref{fig-skew} provides a comprehensive analysis of the relationship among accuracy, model bias, and the data bias level $\alpha$ on the CelebA dataset, focusing on the \texttt{BigNose} recognition task. As $\alpha$ decreases, the deployed models learn more severe biases.
By applying our proposed approach to these diverse deployed models, significant reductions in bias and improvements in accuracy are achieved across all models. Importantly, these improvements are consistent and do not deteriorate as $\alpha$ decreases.

\myparagraph{The probability evaluation functions of deployed models.} \hspace{2mm}
Our method is grounded in Bayesian analysis and, as such, remains fundamentally independent of the probability evaluation functions employed by the deployed model, although our theoretical derivation uses the softmax probability distribution as a representative example.

To confirm this claim, we present the debiasing results of the deployed model using sigmoid as the probability evaluation function and compare them with the results obtained using softmax in Fig.~\ref{fig-softmax}. It is evident that our method consistently reduces bias in both the deployed model using softmax and the deployed model using sigmoid. This finding demonstrates that our approach is independent of the specific choice of probability evaluation function used in the deployed model and can be effectively applied to both sigmoid and softmax distributions.

\input{tables/table5}


\myparagraph{Robustness to Network Architectures.} \hspace{2mm}
To assess the robustness of our method across different deployed model architectures, we evaluated its performance using ResNet-34~\cite{he2016deep} and ViT-B/32~\cite{50650} as the backbones of the deployed model, as reported in Table~\ref{table:modelArch}. The results demonstrate that our method performs consistently across both convolutional networks and transformers, which are two mainstream architectures. This indicates that our approach is not sensitive to the underlying network architecture.

\section{Conclusion}
This paper introduced the Inference-Time Rule Eraser (Eraser), a novel approach to enhancing fairness in AI systems without requiring access to or modification of model weights. Eraser functions by eliminating biased rules during inference through output modification. We also presented a specific implementation that distills biased rules from the deployed model into a supplementary patched model, which removes them during inference. Experimental results validate the effectiveness and superiority of Eraser in mitigating bias.




\bibliographystyle{IEEEtran}
\bibliography{main}
\vspace{-3 mm}

\vfill

\end{document}

%% file: tables/table1.tex
\begin{table*}[t]
\vspace{1mm}

\caption{Results of deployed models before and after integration with the proposed \textit{Eraser} on CelebA (Table~\ref{table:big:sub1} to \ref{table:big:sub3}), UTKFace (Table~\ref{table:big:sub4}), C-MNIST ( Table~\ref{table:big:sub5}) and ImangeNet-B (Table~\ref{table:big:sub6}). For target task performance, we report average group accuracy (in $\%$, $\uparrow$) and Worst group accuracy (in $\%$, $\uparrow$). For model bias, we report Equalodds (in $\%$, $\downarrow$).}
    \label{table:1}
    \renewcommand{\arraystretch}{1.03}
\begin{subtable}[b]{0.497\linewidth}
\centering
\resizebox*{8cm}{!}{
\begin{tabular}{lccc}
\toprule
\multirow{2}{*}{\textbf{CelebA, \textit{BigNose}}}&\multirow{2}{*}{\textbf{ \makecell{ Average \\ACC $\uparrow$}}}&\multirow{2}{*}{\textbf{\makecell{ Worst \\ACC $\uparrow$}}}&\multirow{2}{*}{\textbf{ \makecell{ Model \\Bias $\downarrow$}}}\\
\\
\midrule
{Vanilla model} & {69.48}\tiny{$\pm$0.2} & {36.45}\tiny{$\pm$0.6} & {23.26}\tiny{$\pm$0.3}\\
\rowcolor{mygray}
{Vanilla model  +\textit{  Eraser}}& \textbf{75.14}\tiny{$\pm$0.2} & \textbf{71.52}\tiny{$\pm$0.5} & \textbf{3.72}\tiny{$\pm$0.2}\\
\midrule
{Fair model (AdvDebias)} & {68.41}\tiny{$\pm$0.4}& {36.21}\tiny{$\pm$0.8}& {12.59}\tiny{$\pm$0.5}\\
\rowcolor{mygray}
{Fair model  +\textit{  Eraser}} & \textbf{72.93}\tiny{$\pm$0.4} & \textbf{60.12}\tiny{$\pm$0.5} & \textbf{4.59}\tiny{$\pm$0.4}\\
\midrule
{Fair model (ConDebias)} & {72.96}\tiny{$\pm$0.2} & {53.04}\tiny{$\pm$0.6} & {8.59}\tiny{$\pm$0.5}\\
\rowcolor{mygray}
{Fair model  +\textit{  Eraser}} & \textbf{75.52}\tiny{$\pm$0.4} & \textbf{72.54}\tiny{$\pm$0.5} & \textbf{6.01}\tiny{$\pm$0.4}\\
\midrule
{Unfair model} & {69.89}\tiny{$\pm$1.0} & {42.20}\tiny{$\pm$1.7} & {20.70}\tiny{$\pm$1.4}\\
\rowcolor{mygray}
{Unfair model  +\textit{  Eraser}} & \textbf{72.92}\tiny{$\pm$0.7} & \textbf{66.85}\tiny{$\pm$1.0} & \textbf{3.34}\tiny{$\pm$0.7}\\
\bottomrule
\end{tabular}
}
\setcounter{subtable}{0}
\vspace{0.1cm}
\caption{Results on CelebA when the target label is \emph{BigNose}}
\label{table:big:sub1}
\end{subtable}
\vspace{0.1cm}
\begin{subtable}[b]{0.497\linewidth}
\centering
\resizebox*{8cm}{!}{
\begin{tabular}{lccc}
\toprule
\multirow{2}{*}{\textbf{UTKFace}}&\multirow{2}{*}{\textbf{ \makecell{ Average \\ACC $\uparrow$}}}&\multirow{2}{*}{\textbf{\makecell{ Worst \\ACC $\uparrow$}}}&\multirow{2}{*}{\textbf{ \makecell{ Model \\Bias $\downarrow$}}}\\
\\
\midrule
{Vanilla model} & {86.59}\tiny{$\pm$0.4}  & {72.61}\tiny{$\pm$0.4} & {18.28}\tiny{$\pm$0.3}\\
\rowcolor{mygray}
{Vanilla model  +\textit{  Eraser}} & \textbf{90.09}\tiny{$\pm$0.3}  & \textbf{85.03}\tiny{$\pm$0.4} & \textbf{5.40}\tiny{$\pm$0.3}\\
\midrule
{Fair model (AdvDebias)} & {87.91}\tiny{$\pm$1.2}  & {75.62}\tiny{$\pm$1.9} & {10.82}\tiny{$\pm$1.5}\\
\rowcolor{mygray}
{Fair model  +\textit{  Eraser}} & \textbf{88.85}\tiny{$\pm$1.0} & \textbf{86.03}\tiny{$\pm$1.4} & \textbf{4.36}\tiny{$\pm$1.1}\\
\midrule
{Fair model (ConDebias)} & {86.23}\tiny{$\pm$1.0}  & {76.45}\tiny{$\pm$1.5} & {10.57}\tiny{$\pm$1.3}\\
\rowcolor{mygray}
{Fair model  +\textit{  Eraser}} & \textbf{88.64}\tiny{$\pm$0.6}  & \textbf{85.36}\tiny{$\pm$0.8} & \textbf{6.20}\tiny{$\pm$0.6}\\
\midrule
{Unfair model } & {84.41}\tiny{$\pm$0.7}  & {70.41}\tiny{$\pm$1.1} & {22.14}\tiny{$\pm$0.8}\\
\rowcolor{mygray}
{Unfair model  +\textit{  Eraser}} & \textbf{88.03}\tiny{$\pm$0.5}  & \textbf{82.45}\tiny{$\pm$0.9} & \textbf{3.05}\tiny{$\pm$0.6}\\
\bottomrule
\end{tabular}
}
\setcounter{subtable}{3}
\vspace{0.1cm}
\caption{Results on UTK-face with \emph{ethnicity} as the bias attribute}
\label{table:big:sub4}
\end{subtable}
\vspace{0.1cm}
\begin{subtable}[b]{0.497\linewidth}
\centering
\resizebox*{8cm}{!}{
\begin{tabular}{lccc}
\toprule
\multirow{2}{*}{\textbf{CelebA, \textit{Blonde}}}&\multirow{2}{*}{\textbf{ \makecell{ Average \\ACC $\uparrow$}}}&\multirow{2}{*}{\textbf{\makecell{ Worst \\ACC $\uparrow$}}}&\multirow{2}{*}{\textbf{ \makecell{ Model \\Bias $\downarrow$}}}\\
\\
\midrule
{Vanilla model} & {82.39}\tiny{$\pm$0.3} & {46.13}\tiny{$\pm$0.7} & {22.06}\tiny{$\pm$0.6}\\
\rowcolor{mygray}
{Vanilla model  +\textit{  Eraser}} & \textbf{91.54}\tiny{$\pm$0.3} & \textbf{85.64}\tiny{$\pm$0.5} & \textbf{6.24}\tiny{$\pm$0.3}\\
\midrule
{Fair model (AdvDebias)} & {85.07}\tiny{$\pm$0.5} & {61.72}\tiny{$\pm$0.9} & {11.12}\tiny{$\pm$0.6} \\
\rowcolor{mygray}
{Fair model  +\textit{  Eraser}} & \textbf{90.31}\tiny{$\pm$0.5} & \textbf{85.34}\tiny{$\pm$0.7} & \textbf{5.35}\tiny{$\pm$0.7}\\
\midrule
{Fair model (ConDebias)} & {82.46}\tiny{$\pm$0.3} & {55.05}\tiny{$\pm$0.7} & {13.63}\tiny{$\pm$0.4}\\
\rowcolor{mygray}
{Fair model  +\textit{  Eraser}} & \textbf{90.46}\tiny{$\pm$0.2} & \textbf{80.33}\tiny{$\pm$0.6} & \textbf{5.94}\tiny{$\pm$0.3}\\
\midrule
{Unfair model} & {80.45}\tiny{$\pm$1.3} & {38.93}\tiny{$\pm$1.4} & {25.51}\tiny{$\pm$1.3}\\
\rowcolor{mygray}
{Unfair model  +\textit{  Eraser}} & \textbf{92.21}\tiny{$\pm$0.7} & \textbf{88.35}\tiny{$\pm$0.8} & \textbf{4.71}\tiny{$\pm$0.8}\\
\bottomrule
\end{tabular}
}
\setcounter{subtable}{1}
\vspace{0.1cm}
\caption{Results on CelebA when the target label is \emph{Blonde}}
\label{table:big:sub2}
\end{subtable}
\vspace{0.1cm}
    \begin{subtable}[b]{0.497\linewidth}
    \centering
\resizebox*{8cm}{!}{
\begin{tabular}{lccc}
\toprule
\multirow{2}{*}{\textbf{C-MNIST}}&\multirow{2}{*}{\textbf{ \makecell{ Average \\ACC $\uparrow$}}}&\multirow{2}{*}{\textbf{\makecell{ Worst \\ACC $\uparrow$}}}&\multirow{2}{*}{\textbf{ \makecell{ Model \\Bias $\downarrow$}}}\\
\\
\midrule
{Vanilla model} & {77.90}\tiny{$\pm$0.3} & {12.91}\tiny{$\pm$2.1} & {14.46}\tiny{$\pm$1.3}\\
\rowcolor{mygray}
{Vanilla model  +\textit{  Eraser}} & \textbf{95.88}\tiny{$\pm$0.3} & \textbf{76.35}\tiny{$\pm$1.3} & \textbf{2.50}\tiny{$\pm$0.3}\\
\midrule
{Fair model (AdvDebias)} & {94.35}\tiny{$\pm$0.3} & {19.54}\tiny{$\pm$1.7} & {3.79}\tiny{$\pm$0.8}\\
\rowcolor{mygray}
{Fair model  +\textit{  Eraser}} & \textbf{96.58}\tiny{$\pm$0.4} & \textbf{74.95}\tiny{$\pm$1.1} & \textbf{2.38}\tiny{$\pm$0.3}\\
\midrule
{Fair model (ConDebias)} & {96.32}\tiny{$\pm$0.4}  & {89.12}\tiny{$\pm$2.0} & {2.08}\tiny{$\pm$0.7}\\
\rowcolor{mygray}
{Fair model  +\textit{  Eraser}} & \textbf{96.45}\tiny{$\pm$0.3}  & \textbf{89.88}\tiny{$\pm$1.4} & \textbf{1.77}\tiny{$\pm$0.3}\\
\midrule
{Unfair model} & {73.65}\tiny{$\pm$0.9}  & {11.90}\tiny{$\pm$2.2} & {18.65}\tiny{$\pm$1.4}\\
\rowcolor{mygray}
{Unfair model  +\textit{  Eraser}} & \textbf{94.86}\tiny{$\pm$0.4}  & \textbf{57.08}\tiny{$\pm$0.8} & \textbf{3.20}\tiny{$\pm$0.3}\\
\bottomrule
\end{tabular}
}
\setcounter{subtable}{4}
\vspace{0.1cm}
\caption{Results on C-MNIST with \emph{background color} as the bias attribute }
\label{table:big:sub5}
\end{subtable}
\vspace{0.1cm}
\begin{subtable}[b]{0.497\linewidth}
\centering
\resizebox*{8cm}{!}{
\begin{tabular}{lccc}
\toprule
\multirow{2}{*}{\textbf{CelebA, \textit{Attractive}}}&\multirow{2}{*}{\textbf{ \makecell{ Average \\ACC $\uparrow$}}}&\multirow{2}{*}{\textbf{\makecell{ Worst \\ACC $\uparrow$}}}&\multirow{2}{*}{\textbf{ \makecell{ Model \\Bias $\downarrow$}}}\\
\\
\midrule
{Vanilla model} & {76.85}\tiny{$\pm$0.4} & {66.20}\tiny{$\pm$0.7} & {20.92}\tiny{$\pm$0.4} \\
\rowcolor{mygray}
{Vanilla model  +\textit{  Eraser}} & \textbf{78.76}\tiny{$\pm$0.4} & \textbf{75.73}\tiny{$\pm$0.6} & \textbf{2.30}\tiny{$\pm$0.4} \\
\midrule
{Fair model (AdvDebias)} & {80.25}\tiny{$\pm$0.7} & {74.13}\tiny{$\pm$1.3} & {7.01}\tiny{$\pm$0.9} \\
\rowcolor{mygray}
{Fair model  +\textit{  Eraser}} & \textbf{80.33}\tiny{$\pm$0.5} & \textbf{76.24}\tiny{$\pm$0.7} & \textbf{5.92}\tiny{$\pm$0.7}\\
\midrule
{Fair model (ConDebias)} & {78.82}\tiny{$\pm$0.8} & {72.00}\tiny{$\pm$0.9} & {11.70}\tiny{$\pm$0.8}\\
\rowcolor{mygray}
{Fair model  +\textit{  Eraser}} & \textbf{79.78}\tiny{$\pm$0.5} & \textbf{76.23}\tiny{$\pm$0.7} & \textbf{1.45}\tiny{$\pm$0.5}\\
\midrule
{Unfair model} & {76.45}\tiny{$\pm$1.0} & {64.74}\tiny{$\pm$1.3} & {20.13}\tiny{$\pm$1.1}\\
\rowcolor{mygray}
{Unfair model  +\textit{  Eraser}} & \textbf{78.28}\tiny{$\pm$0.6} & \textbf{77.43}\tiny{$\pm$0.7} & \textbf{1.01}\tiny{$\pm$0.6}\\
\bottomrule
\end{tabular}
}
\setcounter{subtable}{2}
\vspace{0.1cm}
\caption{Results on CelebA when the target label is \emph{Attractive}}
\label{table:big:sub3}
\end{subtable}
\vspace{0.1cm}
\begin{subtable}[b]{0.497\linewidth}
\centering
\resizebox*{8cm}{!}{
\begin{tabular}{lccc}
\toprule
\multirow{2}{*}{\textbf{ImageNet-B}}&\multirow{2}{*}{\textbf{ \makecell{ Average \\ACC $\uparrow$}}}&\multirow{2}{*}{\textbf{\makecell{ Worst \\ACC $\uparrow$}}}&\multirow{2}{*}{\textbf{ \makecell{ Model \\Bias $\downarrow$}}}\\
\\
\midrule
{Vanilla model} & {61.24}\tiny{$\pm$0.3} & {26.01}\tiny{$\pm$0.9} & {13.84}\tiny{$\pm$0.4}\\
\rowcolor{mygray}
{Vanilla model  +\textit{  Eraser}} & \textbf{75.33}\tiny{$\pm$0.3} & \textbf{49.63}\tiny{$\pm$0.5} & \textbf{6.47}\tiny{$\pm$0.3}\\
\midrule
{Fair model (AdvDebias)} & {64.51}\tiny{$\pm$0.6}  & {29.38}\tiny{$\pm$0.8} & {10.62}\tiny{$\pm$0.4}\\
\rowcolor{mygray}
{Fair model  +\textit{  Eraser}} & \textbf{68.48}\tiny{$\pm$0.2}  & \textbf{43.79}\tiny{$\pm$0.5} & \textbf{6.54}\tiny{$\pm$0.2}\\
\midrule
{Fair model (ConDebias)} & {63.26}\tiny{$\pm$0.4}  & {28.03}\tiny{$\pm$1.0} & {10.27}\tiny{$\pm$0.7}\\
\rowcolor{mygray}
{Fair model  +\textit{  Eraser}} & \textbf{67.68}\tiny{$\pm$0.3}  & \textbf{42.05}\tiny{$\pm$0.7} & \textbf{6.96}\tiny{$\pm$0.4}\\
\midrule
{UnFair model} & {56.34}\tiny{$\pm$0.4}  & {18.02}\tiny{$\pm$1.1} & {15.24}\tiny{$\pm$0.3}\\
\rowcolor{mygray}
{UnFair model  +\textit{  Eraser}} & \textbf{71.82}\tiny{$\pm$0.5} & \textbf{46.87}\tiny{$\pm$0.6} & \textbf{6.83}\tiny{$\pm$0.3}\\
\bottomrule
\end{tabular}
}
\setcounter{subtable}{5}
\vspace{0.1cm}
\caption{Results on ImageNet-B with \emph{natural noise} as the bias attribute}
\label{table:big:sub6}
\end{subtable}

\vspace{-22pt}

\end{table*}

%% file: tables/table2.tex
\begin{table*}[t]
\caption{Results of deployed models, both before and after integration with the proposed Eraser, on structured datasets LSAC and COMPAS.}
\label{table:2}
\renewcommand{\arraystretch}{1.2}
\centering
\resizebox{0.7\textwidth}{!}{
\begin{tabular}{l | c c c c c | c c c c}
\toprule
\multirow{3}{*}{Method} &&\multicolumn{3}{c}{\textbf{LSAC}} &&& \multicolumn{3}{c}{\textbf{COMPAS}}\\[1pt]

\cline{3-5} \cline{8-10} 
&& \multirow{2}{*}{\textbf{ \makecell{ Average \\ACC $\uparrow$}}}&\multirow{2}{*}{\textbf{\makecell{ Worst \\ACC $\uparrow$}}}&\multirow{2}{*}{\textbf{ \makecell{ Model \\Bias $\downarrow$}}}
&&&\multirow{2}{*}{\textbf{ \makecell{ Average \\ACC $\uparrow$}}}&\multirow{2}{*}{\textbf{\makecell{ Worst \\ACC $\uparrow$}}}&\multirow{2}{*}{\textbf{\makecell{ Model \\Bias $\downarrow$}}}
\\ &&&&&&&&\\[1pt]

\hline 
\hline
Vanilla model &&64.44 \tiny{$\pm$0.5}&  19.02 \tiny{$\pm$1.3} & 26.48 \tiny{$\pm$0.8} &&&62.35 \tiny{$\pm$0.6}&45.47 \tiny{$\pm$1.5} & 19.45 \tiny{$\pm$0.9}  \\ 
\rowcolor{mygray} Vanilla model+Eraser && \textbf{69.01} \tiny{$\pm$0.6}&\textbf{57.04} \tiny{$\pm$0.9} & \textbf{4.11} \tiny{$\pm$0.7} &&& \textbf{63.59} \tiny{$\pm$0.4} &\textbf{58.43} \tiny{$\pm$0.6} & \textbf{2.03} \tiny{$\pm$0.5}  \\ 

\hline

Fair model (AdvDebias)  &&65.73 \tiny{$\pm$0.9}&  46.35 \tiny{$\pm$1.7} & 10.59 \tiny{$\pm$1.3} &&&54.49 \tiny{$\pm$0.8}&41.72 \tiny{$\pm$1.6} & 14.78 \tiny{$\pm$1.2}  \\ 
\rowcolor{mygray} Fair model+Eraser && \textbf{69.31} \tiny{$\pm$0.6}&\textbf{58.61} \tiny{$\pm$1.0} & \textbf{4.23} \tiny{$\pm$0.7} &&& \textbf{59.45} \tiny{$\pm$0.4} &\textbf{58.32} \tiny{$\pm$0.9} & \textbf{2.87} \tiny{$\pm$0.6}  \\

\hline

Fair model (ConDebias)  &&64.51 \tiny{$\pm$1.2}&  43.82 \tiny{$\pm$1.6} & 12.31 \tiny{$\pm$1.3} &&&59.98 \tiny{$\pm$0.8}&40.45 \tiny{$\pm$1.4} & 4.28 \tiny{$\pm$1.2}  \\ 
\rowcolor{mygray} Fair model+Eraser && \textbf{68.66} \tiny{$\pm$0.6}&\textbf{56.81} \tiny{$\pm$0.7} & \textbf{3.80} \tiny{$\pm$0.6} &&& \textbf{62.97} \tiny{$\pm$0.6} &\textbf{58.31} \tiny{$\pm$0.7} & \textbf{2.65} \tiny{$\pm$0.6}  \\

\hline

Unfair model &&63.60 \tiny{$\pm$1.4}&  17.01 \tiny{$\pm$2.3} & 29.98 \tiny{$\pm$1.9} &&&61.38 \tiny{$\pm$1.2}&31.30 \tiny{$\pm$2.0} & 38.07 \tiny{$\pm$1.5}  \\ 
\rowcolor{mygray} Unfair model+Eraser && \textbf{66.75} \tiny{$\pm$0.7}&\textbf{60.92} \tiny{$\pm$0.8} & \textbf{4.96} \tiny{$\pm$0.7} &&& \textbf{62.92} \tiny{$\pm$0.7} &\textbf{61.82} \tiny{$\pm$0.9} & \textbf{2.01} \tiny{$\pm$0.8}   

\\
\hline

\end{tabular}}

\label{table:celeba}
\end{table*}

%% file: tables/table3.tex
\begin{table*}[!thb]
    \caption{Debiasing results for multi-biases on UrbanCars dataset. We report background (BG) bias, co-occurring object (Co-obj) bias, and the average of BG bias and Co-obj bias.}
    \label{table:urbancar}
    \renewcommand{\arraystretch}{1.35}
    \centering
    \resizebox{0.75\textwidth}{!}{
    \begin{tabular}{c l | c c | c c c}
    \hline
        \multirow{2}{*}{ } & \multirow{2}{*}{Method} & \multicolumn{2}{c|}{\textbf{Accuracy}} & \multicolumn{3}{c}{\textbf{Model Bias}} \\[1pt]
        \cline{3-4} \cline{5-7}
        & &  \textbf{Average ACC$\uparrow$} & \textbf{Worst ACC$\uparrow$} & \textbf{BG Bias$\downarrow$} & \textbf{Co-obj Bias$\downarrow$} & \textbf{Avg Bias$\downarrow$} \\[2pt] 
        \hline
        \hline
        & Vanilla model & 77.27 \tiny{$\pm$0.6} & 28.05 \tiny{$\pm$1.4} & 32.07 \tiny{$\pm$1.3} & 28.01 \tiny{$\pm$2.2} & 30.04 \tiny{$\pm$1.6}  \\
        \rowcolor{mygray}
        \cellcolor{white}& Vanilla model + Eraser$_\text{\,BG + Co-obj}$   & \textbf{86.63} \tiny{$\pm$0.7} & \textbf{77.82} \tiny{$\pm$1.1} & \textbf{4.84} \tiny{$\pm$1.2} & \textbf{3.25} \tiny{$\pm$0.9} & \textbf{4.05} \tiny{$\pm$0.6} 
        \\
        \cline{2-7}
        & Fair model (AdvDebias)  & 82.13 \tiny{$\pm$1.3} & 62.42 \tiny{$\pm$2.5} & 19.36 \tiny{$\pm$1.7} & 10.77 \tiny{$\pm$1.9} & 15.06 \tiny{$\pm$1.8}  \\
        \rowcolor{mygray}
        \cellcolor{white} & Fair model + Eraser$_\text{\,BG + Co-obj}$  & \textbf{86.44} \tiny{$\pm$0.9} & \textbf{76.56} \tiny{$\pm$2.1} & \textbf{4.70} \tiny{$\pm$0.6} & \textbf{3.12} \tiny{$\pm$0.7} & \textbf{3.91} \tiny{$\pm$0.7} \\
                \cline{2-7}
        & Fair model (ConDebias)  & 84.74 \tiny{$\pm$1.1} & 67.25 \tiny{$\pm$3.4} & 20.64 \tiny{$\pm$1.8} & 8.56 \tiny{$\pm$1.6} & 14.6 \tiny{$\pm$1.7}  \\
        \rowcolor{mygray}
        \cellcolor{white} & Fair model + Eraser$_\text{\,BG + Co-obj}$  & \textbf{87.39} \tiny{$\pm$0.7} & \textbf{77.62} \tiny{$\pm$1.2} & \textbf{5.02} \tiny{$\pm$1.4} & \textbf{4.64} \tiny{$\pm$0.8} & \textbf{4.83} \tiny{$\pm$1.2} \\
                \cline{2-7}

        & Unfair model  & 75.26 \tiny{$\pm$1.6} & 17.62 \tiny{$\pm$3.1} &33.67  \tiny{$\pm$1.1} & 38.45 \tiny{$\pm$2.9} & 36.06 \tiny{$\pm$1.9}  \\
        \rowcolor{mygray}
        \cellcolor{white}\multirow{-8}{*}{\makecell[c] { \cellcolor{white} Car\\ \cellcolor{white} Object}} & Unfair model + Eraser$_\text{\,BG + Co-obj}$  & \textbf{85.40} \tiny{$\pm$0.9} & \textbf{67.26} \tiny{$\pm$1.3} & \textbf{1.20 \tiny{$\pm$1.4}} & \textbf{4.38} \tiny{$\pm$0.3} & \textbf{2.79} \tiny{$\pm$0.9} \\
    \bottomrule
\end{tabular}}
\end{table*}

%% file: tables/table4.tex
\begin{table*}[t]
\centering
\caption{Debiasing results of different post-training debiasing methods on six tasks under laboratory conditions. 
The best results are highlighted in \textbf{bold}. The second-best results are \underline{underlined}. 
The Equalodds and CalEqualodds are not suitable for multi-classification tasks (indicated by  ``-'').}
\label{laboratory}
\renewcommand\arraystretch{1.8}
\resizebox{0.94\textwidth}{!}{
\setlength{\tabcolsep}{0.1mm}{
\begin{tabular}{l |c:c:c|c:c:c|c:c:c|c:c:c|c:c:c|c:c:c} 
\toprule[1pt]
 \rowcolor{white} &  \multicolumn{3}{c|}{\textbf{CelebA,\textit{BigNose}}} & \multicolumn{3}{c|}{\textbf{CelebA,\textit{Blonde}}} & \multicolumn{3}{c|}{\textbf{CelebA,\textit{Attractive}}}& \multicolumn{3}{c|}{\textbf{UTKFace}}& \multicolumn{3}{c|}{\textbf{C-MNIST}}& \multicolumn{3}{c}{\textbf{ImageNe-B}} \\ \cline{2-19} 
 {\textbf{Methods}} & \multirow{2}{*}{\textbf{\makecell{Avg. \\[5pt] ACC $\uparrow$}}} & \multirow{2}{*}{\textbf{\makecell{ Worst \\[5pt] ACC $\uparrow$}}} & \multirow{2}{*}{\textbf{\makecell{ Model \\[5pt] Bias $\downarrow$}}} 
 & \multirow{2}{*}{\textbf{\makecell{ Avg. \\[5pt] ACC $\uparrow$}}} & \multirow{2}{*}{\textbf{\makecell{ Worst \\[5pt] ACC $\uparrow$}}} & \multirow{2}{*}{\textbf{\makecell{ Model \\[5pt] Bias $\downarrow$}}} 
 & \multirow{2}{*}{\textbf{\makecell{ Avg. \\[5pt] ACC $\uparrow$}}} & \multirow{2}{*}{\textbf{\makecell{ Worst \\[5pt] ACC $\uparrow$}}} & \multirow{2}{*}{\textbf{\makecell{ Model \\[5pt] Bias $\downarrow$}}} 
 & \multirow{2}{*}{\textbf{\makecell{ Avg. \\[5pt] ACC $\uparrow$}}} & \multirow{2}{*}{\textbf{\makecell{ Worst \\[5pt] ACC $\uparrow$}}} & \multirow{2}{*}{\textbf{\makecell{ Model \\[5pt] Bias $\downarrow$}}} 
 & \multirow{2}{*}{\textbf{\makecell{ Avg. \\[5pt] ACC $\uparrow$}}} & \multirow{2}{*}{\textbf{\makecell{ Worst \\[5pt] ACC $\uparrow$}}} & \multirow{2}{*}{\textbf{\makecell{ Model \\[5pt] Bias $\downarrow$}}} 
 & \multirow{2}{*}{\textbf{\makecell{ Avg. \\[5pt] ACC $\uparrow$}}} & \multirow{2}{*}{\textbf{\makecell{ Worst \\[5pt] ACC $\uparrow$}}} & \multirow{2}{*}{\textbf{\makecell{ Model \\[5pt] Bias $\downarrow$}}} 
 \\&&&&&&&&&&&&&&&&& 
 \\
\hline 
\emph{Vanilla}                      & $69.48$ & $36.45$ & $23.26$  & $82.39$ & $46.13$ & $22.06$ & $76.85$ & $66.20$ & $20.92$& $86.59$ & $72.61$ & $18.28$ & $77.90$ & $12.91$ & $14.46$ & $61.24$ & $26.01$ & $13.84$ \\
\hline
\emph{Equalodds}                      &$64.95$ & $42.23$ & $\bf{1.50}$  & $71.86$ & $45.12$ & $\bf{0.70}$ &$71.74$ & $\underline{71.21}$ & $\bf{1.06}$  & $80.59$ & $73.14$ & $\bf{1.67}$ & $-$ & $-$ & $-$ & $-$ & $-$ & $-$ \\

\emph{CalEqualodds}    & $57.87$ & $36.45$ & $20.62$ & $68.48$ & $29.36$ & $8.62$ &$77.51$ & $69.20$ & $16.49$  & $76.08$ & $42.43$ & $17.39$ & $-$ & $-$ & $-$ & $-$ & $-$ & $-$ \\
\emph{FAAP}     &  $\underline{72.82}$ & $46.74$ & $6.97$& $86.98$ & $66.45$ & $5.24$ & $77.76$ & $67.23$ & $8.26$ & $\underline{88.12}$ & $\underline{78.13}$ & $9.27$ & $84.67$ & $\underline{36.57}$ & $\underline{4.95}$ & $\underline{73.67}$ & $\underline{44.29}$ & $7.89$ \\
\emph{Repro}     & $70.35$ & $47.53$ & $9.23$  & $84.49$ & $60.04$ & $11.04$ &$78.04$ & $64.31$ & $11.13$ & $87.12$ & $77.62$ & $10.28$ & $81.49$ & $25.17$ & $8.27$ & $72.51$ & $42.31$ & $9.57$ \\
\emph{CARE}      & $71.53$ & $46.96$ & $19.79$ & $86.97$ & $77.59$ & $13.58$ & $\underline{78.53}$ & $68.79$ & $19.87$ & $87.61$ & $77.79$ & $13.92$ & $86.07$ & $20.45$ & $7.90$ & $72.59$ & $41.63$ & $7.56$ \\
\emph{DFR}                & $70.15$ & $\underline{47.64}$ & $20.83$ & $\underline{89.49}$ & $\underline{78.35}$ & $12.25$ &$77.61$ & $68.39$ & $18.34$  & $87.98$ & $78.05$ & $11.05$ & $\underline{86.43}$ & $21.56$ & $7.61$ & $72.62$ & $42.85$ & $\underline{7.39}$ \\
\hline
\emph{Eraser}         &$\bf{75.14}$  & $\bf{71.52}$ & $\underline{3.72}$  & $\bf{91.54}$ & $\bf{85.64}$ & $\underline{6.24}$ & $\bf{78.76}$  & $\bf{75.73}$ & $\underline{2.30}$ & $\bf{90.09}$ & $\bf{85.03}$ & $\underline{5.40}$ & $\bf{95.88}$  & $\bf{76.35}$ & $\bf{2.50}$ & $\bf{75.33}$ & $\bf{49.63}$ & $\bf{6.47}$ \\
\bottomrule[1pt]
\end{tabular}}}
\vspace{-10pt}
\end{table*}

%% file: tables/table5.tex
\begin{table}[t!]

   \caption{Effect of backbone architectures in BigNose on CelebA dataset.}

\resizebox{0.49\textwidth}{!}{
\begin{tabular}{ccccc} \toprule
Backbone & Method & Avg. Acc ($\uparrow$) & Worst Acc ($\uparrow$) & Model Bias ($\downarrow$) \\ \cmidrule[0.5pt]{1-5} \morecmidrules\cmidrule[0.5pt]{1-5}
\multirow{2}{*}{\texttt{Resnet-34}} & \textit{Vanilla} & 69.48$_{\pm 0.2}$ & 36.45$_{\pm 0.6}$& 23.26$_{\pm 0.3}$ \\ \cmidrule[0.3pt]{2-5}
 & \textit{Eraser} & \textbf{75.14}$_{\pm 0.2}$ & \textbf{71.52}$_{\pm 0.5}$& \textbf{3.72}$_{\pm 0.2}$ \\ \cmidrule[0.3pt]{1-5}
\multirow{2}{*}{\texttt{VIT-B/32}} &  \textit{Vanilla} & 60.01$_{\pm 0.1}$ & 11.75$_{\pm 0.6}$ & 27.53$_{\pm 0.3}$\\ \cmidrule[0.3pt]{2-5}
 & \textit{Eraser}& \textbf{66.85}$_{\pm 0.1}$ & \textbf{63.91}$_{\pm 0.4}$ & \textbf{3.75}$_{\pm 0.2}$\\ \bottomrule
\end{tabular}
}
 	
%
\label{table:modelArch}
\end{table}